\documentclass[11pt]{article}
\usepackage{arxiv}
\usepackage[utf8]{inputenc} 
\usepackage[T1]{fontenc}    
\usepackage{hyperref}       
\usepackage{url}            
\usepackage{booktabs}       
\usepackage{amsfonts}       
\usepackage{nicefrac}       
\usepackage{microtype}      
\usepackage{lipsum}
\usepackage{graphicx}
\usepackage{amsmath}
\usepackage{amssymb}
\usepackage{defs}
\usepackage{todonotes}
\usepackage{wrapfig}
\usepackage{graphicx}
\usepackage{nicefrac}
\usepackage[font=small]{caption}
\usepackage[font=small, skip=-1pt]{subcaption}
\usepackage{xcolor}
\usepackage[toc,page,header]{appendix}
\usepackage{minitoc}

\usepackage{xcolor}         
\usepackage{enumitem}
\usepackage[ruled, vlined, linesnumbered]{algorithm2e}%
\usepackage{algpseudocode}
\SetAlFnt{\scriptsize}
\usepackage{rotating}
\usepackage{tabularx}
\usepackage{multirow}
\usepackage{adjustbox}
\newcommand{\model}{\textsc{Automata}}
\newcommand{\figref}[1]{Figure~\ref{#1}}

\newcommand{\secref}[1]{Section~\ref{#1}}

\renewcommand{\eqref}[1]{Equation~(\ref{#1})}

\title{\model\ : Gradient Based Data Subset Selection for Compute-Efficient Hyper-parameter Tuning}

\author{Krishnateja Killamsetty\textsuperscript{1}, \quad  Guttu Sai Abhishek\textsuperscript{2}, \quad  Aakriti\textsuperscript{2}, \quad  Alexandre V. Evfimievski\textsuperscript{3} \\
        \textbf{Lucian Popa\textsuperscript{3}, \quad Ganesh Ramakrishnan\textsuperscript{2}, \quad Rishabh Iyer\textsuperscript{1}}\\
    \textsuperscript{1} The University of Texas at Dallas  \\
    \textsuperscript{2}Indian Institute of Technology Bombay, India \\ 
    \textsuperscript{3} IBM Research\\
    \texttt{\{krishnateja.killamsetty, rishabh.iyer\}@utdallas.edu}\\
    \texttt{\{gsaiabhishek, aakriti, ganesh\}@cse.iitb.ac.in}\\
    \texttt{\{evfimi, lpopa\}@us.ibm.com}\\
}

\begin{document}

\maketitle

\begin{abstract}
Deep neural networks have seen great success in recent years; however, training a deep model is often challenging as its performance heavily depends on the hyper-parameters used. In addition, finding the optimal hyper-parameter configuration, even with state-of-the-art (SOTA) hyper-parameter optimization (HPO) algorithms, can be time-consuming, requiring multiple training runs over the entire dataset for different possible sets of hyper-parameters. Our central insight is that using an informative subset of the dataset for model training runs involved in hyper-parameter optimization, allows us to find the optimal hyper-parameter configuration significantly faster. In this work, we propose \model, a gradient-based subset selection framework for hyper-parameter tuning. We empirically evaluate the effectiveness of \model\ in hyper-parameter tuning through several experiments on real-world datasets in the text, vision, and tabular domains. Our experiments show that using gradient-based data subsets for hyper-parameter tuning achieves significantly faster turnaround times and speedups of \textbf{3$\times$-30$\times$} while achieving comparable performance to the hyper-parameters found using the entire dataset. 

\end{abstract}

\section{Introduction}
\label{sec:intro}

In recent years, deep learning systems have found great success in a wide range of tasks, such as object recognition~\cite{He2015DelvingDI}, speech recognition~\cite{hersheymultitalker}, and machine translation~\cite{barrault-etal-2019-findings}, making people's lives easier on a daily basis. However, in the quest for near-human performance, more complex and deeper machine learning models trained on increasingly large datasets are being used at the expense of substantial computational costs. Furthermore, deep learning is associated with a significantly large number of hyper-parameters such as the learning algorithm, batch size, learning rate, and model configuration parameters ({\em e.g.}, depth, number of hidden layers, {\em etc.}) that need to be tuned. Hence, running extensive hyper-parameter tuning and auto-ml pipelines is becoming increasingly necessary to achieve state-of-the-art models. However, tuning the hyper-parameters requires multiple training runs over the entire datasets (which are significantly large nowadays), resulting in staggering compute costs, running times, and, more importantly, CO2 emissions.
\begin{wrapfigure}{L}{0.4\textwidth}
    \centering
    \includegraphics[width=6cm, height=4cm]{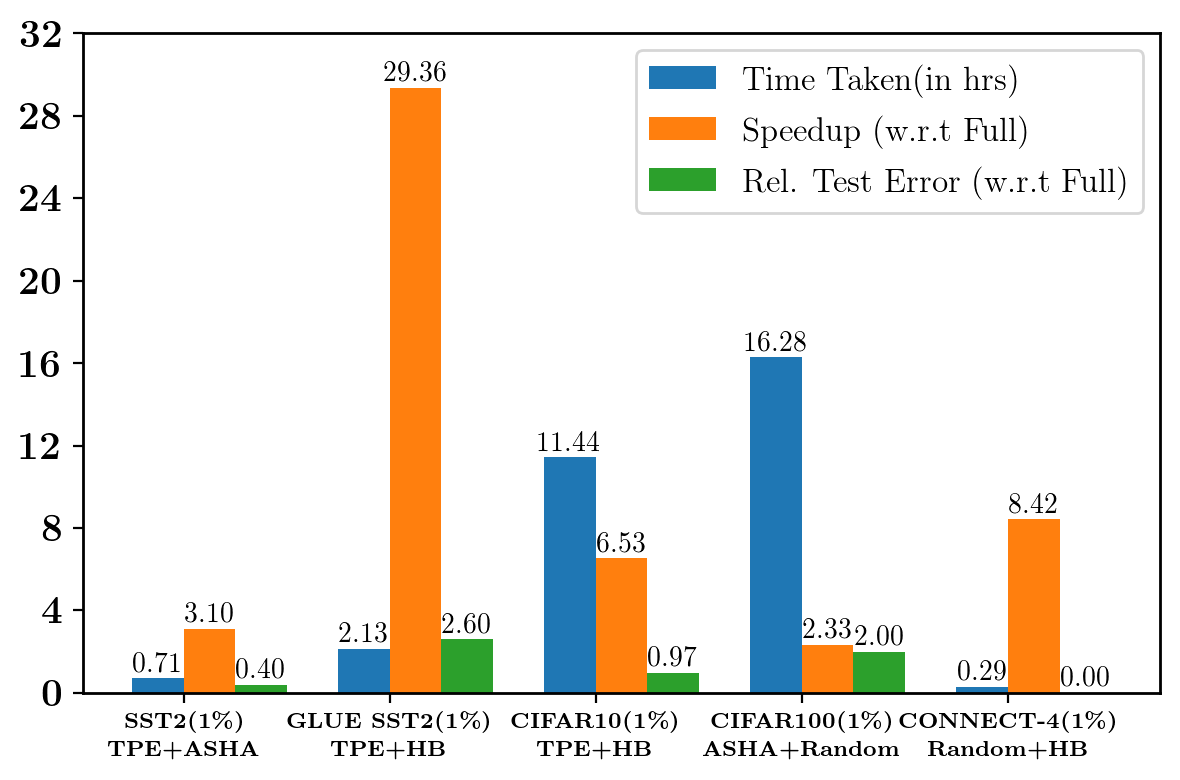}
    \caption{\scriptsize \model{}'s performance summary showing speedups, relative test errors, and tuning times on SST2, glue-SST2, CIFAR10, CIFAR100, and CONNECT-4 datasets. We observe that \model\ achieves speedups(and similar energy savings) of 10x - 30x with around 2\% performance loss using Hyperband as a scheduler. Similarly, even when using a more efficient ASHA scheduler, \model\ achieves a speedup of around 2x-3x with a performance loss of 0\%-2\%.}
    \label{fig:perf_summary}
\end{wrapfigure}


To give an idea of staggering compute costs, we consider an image classification task on a relatively simple CIFAR-10 dataset where a single training run using a relatively simple model class of Residual Networks~\cite{He2016DeepRL} on a V100 GPU takes around 6 hours. If we perform 1000 training runs (which is not uncommon today) naively using grid search for hyper-parameter tuning, it will take 6000 GPU hours. The resulting CO2 emissions would be between 640 to 1400 kg of CO2 emitted\footnote{\url{https://mlco2.github.io/impact/\#compute}}, which is equivalent to 1600 to  3500 miles of car travel in the US. Similarly, the costs of training state-of-the-art NLP models and vision models on larger datasets like ImageNet are even more staggering~\cite{strubell2019energy}\footnote{\url{https://tinyurl.com/a66fexc7}}. 

Naive hyper-parameter tuning methods like grid search~\cite{JMLR:v13:bergstra12a} often fail to scale up with the dimensionality of the search space and are computationally expensive. Hence, more efficient and sophisticated Bayesian optimization methods~\cite{tpe, smbo, pmlr-v28-bergstra13, scalablebayesian, fabolas} have dominated the field of hyper-parameter optimization in recent years. Bayesian optimization methods aim to identify good hyper-parameter configurations quickly by building a posterior distribution over the search space and by adaptively selecting configurations based on the probability distribution. More recent methods~\cite{NIPS2013_f33ba15e, DBLP:journals/corr/SwerskySA14, 10.5555/2832581.2832731, fabolas} try to speed up configuration evaluations for efficient hyper-parameter search; these approaches speed up the configuration evaluation by adaptively allocating more resources to promising hyper-parameter configurations while eliminating poor ones quickly. Existing SOTA methods like SHA~\cite{sha}, Hyperband~\cite{hyperband}, ASHA~\cite{asha} use aggressive early-stopping strategies to stop not-so-promising configurations quickly while allocating more resources to the promising ones. Generally, these resources can be the size of the training set, number of gradient descent iterations, training time, etc. Other approaches like 
~\cite{nickson2014automated,kreuger2015a} try to quickly evaluate a configuration's performance on a large dataset by evaluating the training runs on small, random subsets; they empirically show that small data subsets could suffice to estimate a configuration's quality.

The past works~\cite{nickson2014automated, kreuger2015a} show that very small data subsets can be effectively used to find the best hyper-parameters quickly. However, all these approaches have naively used random training data subsets and did not place much focus on selecting informative subsets instead. Our central insight is that using small informative data subsets allows us to find good hyper-parameter configurations more effectively 
than random data subsets. In this work, we study the application of the gradient-based subset selection approach for the task of hyper-parameter tuning and automatic machine learning. On that note, the use of gradient-based data subset selection approach in supervised learning setting was explored earlier in \textsc{Grad-match}~\cite{killamsetty2021grad} where the authors showed that training on gradient based data subsets allows the models to achieve comparable accuracy to full data training while being significantly faster. 


In this work, we empirically study the advantage of using informative gradient-based subset selection algorithms for the hyper-parameter tuning task and study its accuracy when compared to using random subsets and a full dataset. So essentially, we use subsets of data to tune the hyper-parameters. Once we obtain the tuned hyper-parameters, we then train the model (with the obtained hyper-parameters) on the full dataset. The smaller the data subset we use, the more the speed up and energy savings (and hence the decrease in CO2 emissions). In light of all these insights, we propose \model, an efficient hyper-parameter tuning framework that combines existing hyper-parameter search and scheduling algorithms with intelligent subset selection. We further empirically show the effectiveness and efficiency of \model{} for hyper-parameter tuning, when used with existing hyper-parameter search approaches (more specifically, TPE~\cite{tpe}, Random search~\cite{randomsearch}), and hyper-parameter tuning schedulers (more specifically, Hyperband~\cite{hyperband}, ASHA~\cite{asha}) on datasets spanning text, image, and tabular domains.


\subsection{Related Work}
\noindent \textbf{Hyper-parameter tuning and auto-ml approaches: } A number of algorithms have been proposed for hyper-parameter tuning including grid search\footnote{\url{https://tinyurl.com/3hb2hans}}, bayesian algorithms~\cite{bergstra2011algorithms}, random search~\cite{randomsearch}, etc. Furthermore, a number of scalable toolkits and platforms for hyper-parameter tuning exist like Ray-tune~\cite{liaw2018tune}\footnote{\url{https://docs.ray.io/en/master/tune/index.html}}, H2O automl~\cite{ledell2020h2o}, etc. See~\cite{smith2018disciplined,yu2020hyper} for a survey of current approaches and also tricks for hyper-parameter tuning for deep models. The biggest challenges of existing hyper-parameter tuning approaches are a) the large search space and high dimensionality of hyper-parameters and b) the increased training times of training models. Recent work~\cite{li2018system} has proposed an efficient approach for parallelizing hyper-parameter tuning using Asynchronous Successive Halving Algorithm (ASHA). \model{} is complementary to such approaches and we show that our work can be be combined effectively with them.

\noindent \textbf{Data Subset Selection: } Several recent papers have used submodular functions\footnote{Let $V = \{1, 2, \cdots, n\}$ denote a ground set of items. A set function $f: 2^V \rightarrow \mathbf{R}$ is a submodular ~\cite{fujishige2005submodular} if it satisfies the diminishing returns property: for subsets $S \subseteq T \subseteq V$ and $j \in V \backslash T, f(j | S) \triangleq  f(S \cup j) - f(S) \geq f(j | T)$.} for data subset selection towards various applications like speech recognition~\cite{wei2014unsupervised,wei2014submodular}, machine translation~\cite{kirchhoff2014submodularity} and computer vision~\cite{kaushal2019learning}. Other common approaches for subset selection include the usage of coresets. Coresets are weighted subsets of the data, which approximate certain desirable characteristics of the full data ({\em, e.g.}, the loss function)~\cite{feldman2020core}. Coreset algorithms have been used for several problems including $k$-means and $k$-median clustering~\cite{har2004coresets}, SVMs~\cite{clarkson2010coresets} and Bayesian inference~\cite{campbell2018bayesian}. Recent coreset selection-based methods~\cite{mirzasoleiman2020coresets, killamsetty2021glister, killamsetty2021grad, killamsetty2021retrieve} have shown great promise for efficient and robust training of deep models. \textsc{Craig}~\cite{mirzasoleiman2020coresets} tries to select a coreset summary of the training data that estimate the full training gradient closely. Whereas GLISTER~\cite{killamsetty2021glister} poses the coreset selection problem as a discrete-continuous bilevel optimization problem that minimizes the validation set loss. Similarly, RETRIEVE~\cite{killamsetty2021retrieve} also uses a discrete bilevel coreset selection problem to select unlabeled data subsets for efficient semi-supervised learning. Another approach \textsc{Grad-Match}~\cite{killamsetty2021grad} selects coreset summary that approximately matches the full training loss gradient using orthogonal matching pursuit. 


\subsection{Contributions of the Work}
The contributions of our work can be summarized as follows:

\noindent \textbf{\model{} Framework: } We propose \model\, a framework that combines intelligent subset selection with hyper-parameter search and scheduling algorithms to enable faster hyper-parameter tuning. To our knowledge, ours is the first work that studies the role of intelligent data subset selection for hyper-parameter tuning. In particular, we seek to answer the following question: \emph{Is it possible to use small informative data subsets between 1\% to 30\% for faster configuration evaluations in hyper-parameter tuning, thereby enabling faster tuning times while maintaining comparable accuracy to tuning hyper-parameters on the full dataset?}

\noindent \textbf{Effectiveness of \model\ : } We empirically demonstrate the effectiveness of \model\ framework used in conjunction with existing hyper-parameter search algorithms like TPE, Random Search, and hyper-parameter scheduling algorithms like Hyperband, ASHA through a set of extensive experiments on multiple real-world datasets. We give a summary of the speedup vs. relative performance achieved by \model\ compared to full data training in \figref{fig:perf_summary}. More specifically, \model\ achieves a speedup of 3x - 30x with minimal performance loss for hyper-parameter tuning. Further, in \secref{sec:experiments}, we show that the gradient-based subset selection approach of \model\ outperforms the previously considered random subset selection for hyper-parameter tuning.

\begin{figure*}
    \centering
    \includegraphics[width=14cm, height=7cm]{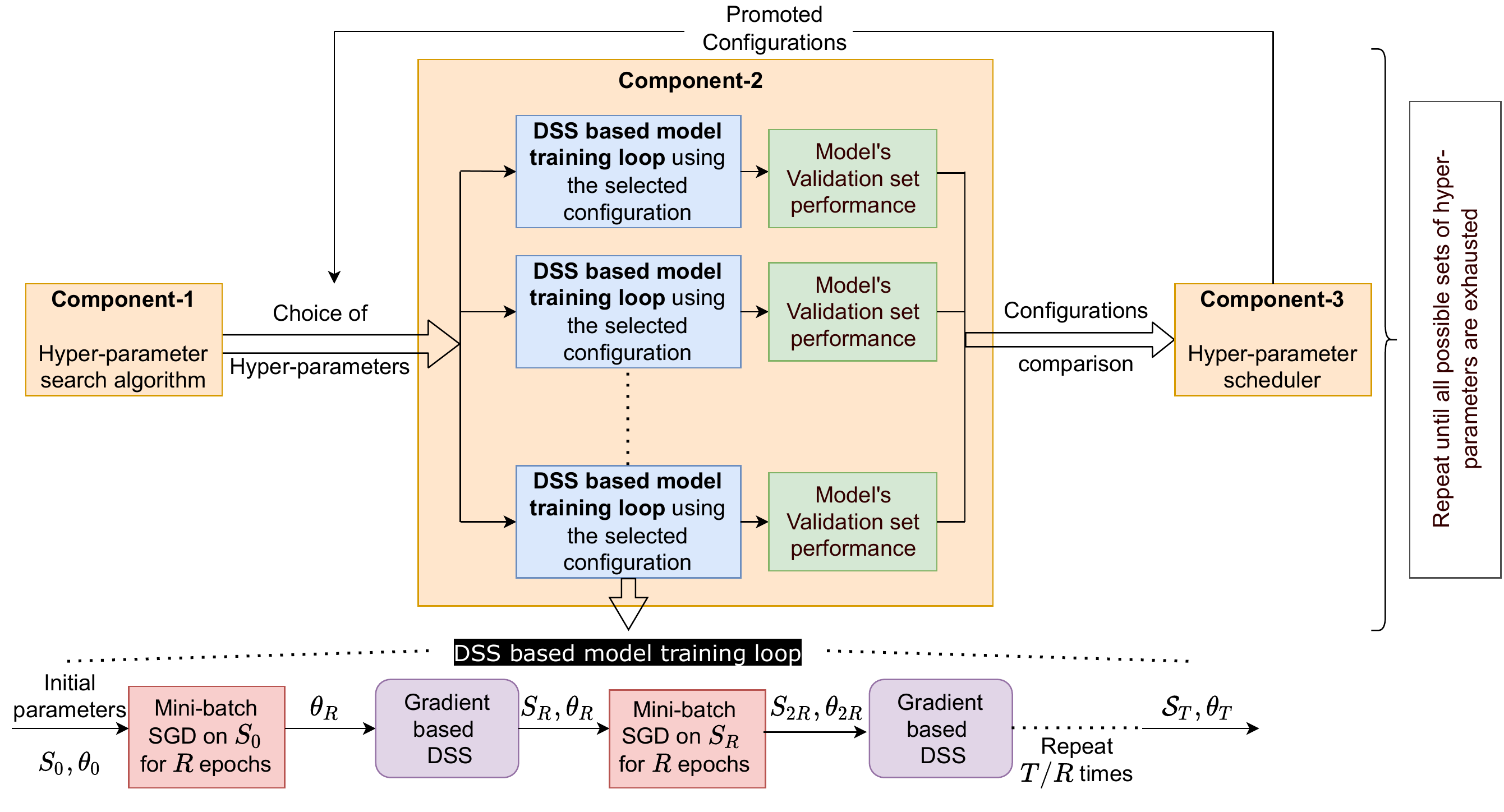}
    \caption{Diagram of \model{}, including hyper-parameter search, subset based configuration evaluation (where models are trained on subsets of data), and hyper-parameter scheduler.}
    
    \label{fig:pipeline}
\end{figure*}

\section{\model\ Framework}
In this section, we present \model{}, a hyper-parameter tuning framework, and discuss its different components shown in \figref{fig:pipeline}. The \model\ framework consists of three components: a hyper-parameter search algorithm that identifies which configuration sets need to be evaluated, a gradient-based subset selection algorithm for training and evaluating each configuration efficiently, and a hyper-parameter scheduling algorithm that provides early stopping by eliminating the poor configurations quickly. With \model{} framework, one can use any of the existing hyper-parameter search and hyper-parameter scheduling algorithms and still achieve significant speedups with minimal performance degradation due to faster configuration evaluation using gradient-based subset training.

\subsection{Notation}
Denote by $\HH$,  the set of configurations selected by the hyper-parameter search algorithm. Let $\Dcal = \{(x_i, y_i)\}_{i=1}^{N}$, denote the set of training examples, and $\Vcal = \{(x_j, y_j)\}_{j=1}^{M}$  the validation set. Let $\theta_i$ denote the classifier model parameters trained using the configuration $i \in \HH$. Let $\Scal_i$ be the subset used for training the $i^{th}$ configuration model $\theta_i$ and $\wb_i$ be its associated weight vector i.e., each data sample in the subset has an associated weight that is used for computing the weighted loss. We superscript the changing variables like model parameters $\theta$, subset $\Scal$ with the timestep $t$ to denote their specific values at that timestep. Next, denote by $L_T^j(\theta_i) = L_T(x_j, y_j, \theta_i)$,  the training loss of the $j^{th}$ data sample in the dataset for $i^{th}$ classifier model, and let $L_T(\theta_i) = \sum_{k \in \Dcal}L_T^k(\theta_i)$ be the loss over the entire training set for $i^{th}$ configuration model. Let, $L_T^j(\Scal, \theta_i) = \sum_{k \in \Xcal} L_T(x_k, y_k, \theta_i)$ be the loss on a subset $\Scal \subseteq \Vcal$ of the training examples at timestep $j$. Let the validation loss be denoted by $L_V$.

\subsection{Component-1: Hyper-parameter Search Algorithm}
Given a hyper-parameter search space, hyper-parameter search algorithms provide a set of configurations that need to be evaluated. A naive way of performing the hyper-parameter search is Grid-Search, which defines the search space as a grid and exhaustively evaluates each grid configuration. However, Grid-Search is a time-consuming process, meaning that thousands to millions of configurations would need to be evaluated if the hyper-parameter space is large. In order to find optimal hyper-parameter settings quickly, Bayesian optimization-based hyper-parameter search algorithms have been developed. To investigate the effectiveness  of \model\ across the spectrum of search algorithms, we used the Random Search method and the Bayesian optimization-based TPE method 
as representative hyper-parameter search algorithms. We provide more details on Random Search and TPE in Appendix~\ref{app:hyperparamsearch}.

\subsection{Component-2: Subset based Configuration Evaluation}
Earlier, we discussed how a hyper-parameter search algorithm presents a set of potential hyper-parameter configurations that need to be evaluated when tuning hyper-parameters. Every time a configuration needs to be evaluated, prior work trained the model on the entire dataset until the resource allocated by the hyper-parameter scheduler is exhausted. Rather than using the entire dataset for training, we propose using subsets of informative data selected based on gradients instead. As a result, given any hyper-parameter search algorithm, we can use the data subset selection to speed up each training epoch by a significant factor \emph{(say 10x)}, thus improving the overall turnaround time of the hyper-parameter tuning. However, the critical advantage of \model\ is that we can achieve speedups while still retaining the hyper-parameter tuning algorithm's performance in finding the best hyper-parameters. The fundamental feature of \model\ is that the subset selected by \model\ changes adaptively over time, based on the classifier model training. Thus, instead of selecting a common subset among all configurations, \model\ selects the subset that best suits each configuration. We give a detailed overview of the gradient-based subset selection process of \model\ below.

\subsubsection{Gradient Based Subset Selection~(GSS)}
The key idea of gradient-based subset selection of \model\ is to select a subset $\Scal$ and its associated weight vector $\wb$ such that the weighted subset loss gradient best approximates the entire training loss gradient. The subset selection of \model\  for $i^{th}$ configuration at time step $t$ is as follows:

\begin{equation}\label{eq:problem_formulation}
 \wb^t_i, \Scal^t_i = \underset{\wb^t_i, \Scal^t_i: |\Scal^t_i| \leq k, \wb^t_i \geq 0}{\argmin} \Vert \sum_{l \in \Scal^t_i} \wb^t_{il} \nabla_{\theta}L_T^l(\theta^t_i) -  \nabla_{\theta}L_T(\theta^t_i)\Vert  + \lambda \left \Vert \wb_i^t \right \Vert ^2
\end{equation}

The additional regularization term prevents assignment of very large weight values to data samples, thereby reducing the possibility of overfitting on a few data samples. A similar formulation for subset selection in the context of efficient supervised learning was employed in a recent work called \textsc{Grad-Match}~\cite{killamsetty2021grad}. The authors of the work~\cite{killamsetty2021grad} proved that the optimization problem given in \eqref{eq:problem_formulation} is approximately submodular. Therefore, the above optimization problem can be solved using greedy algorithms with approximation guarantees~\cite{das2011submodular, minoux1978accelerated}. Similar to ~\textsc{Grad-Match}, we use a greedy algorithm called orthogonal matching pursuit (OMP) to solve the above optimization problem. The goal of \model\ is to accelerate the hyper-parameter tuning algorithm while preserving its original performance. Efficiency is, therefore, an essential factor that \model\ considers even when selecting subsets. Due to this, we employ a faster per-batch subset selection introduced in the work~\cite{killamsetty2021grad} in our experiments, which is described in the following section.

\noindent \textbf{Per-Batch Subset Selection: }
Instead of selecting a subset of data points, one selects a subset of mini-batches by matching the weighted sum of mini-batch training gradients to the full training loss gradients. Therefore, one will have a subset of slected mini-batches and the associated mini-batch weights. One trains the model on the selected mini-batches by performing mini-batch gradient descent using the weighted mini-batch loss. Let us denote the batch size as $B$, and the total number of mini-batches as $b_N = \frac{N}{B}$, and the training set of mini-batches as $\Dcal_{\Bcal}$. Let us denote the number of mini-batches that needs to be selected as $b_k = \frac{k}{B}$. Let us denote the subset of mini-batches that needs to be selected as $\Scal_{\Bcal i}$ and denote the weights associated with mini-batches as $\wb_{\Bcal i} = \{\wb_{\Bcal i1}, \wb_{\Bcal i2} \cdots \wb_{\Bcal ik}\}$ for the $i^{th}$ model configuration. Let us denote the mini-batch gradients as $\nabla_{\theta} L_T^{\Bcal_{i}}(\theta_i), \cdots, \nabla_{\theta} L_T^{\Bcal_{b_N}}(\theta_i)$ be the mini-batch gradients for the $i^{th}$ model configuration.  Let us denote $L_T^{\Bcal}(\theta_i) = \sum_{i \in [1, b_N]}L_T^{\Bcal_{k}}(\theta_i)$ be the loss over the entire training set.

The subset selection problem of mini-batches at time step $t$ can be written as follows:

\begin{equation} \label{eq:perbatch_formulation}
    \wb^t_{\Bcal i}, \Scal^t_{\Bcal i} = \underset{\wb^t_{\Bcal i}, \Scal^t_{\Bcal i}: |\Scal^t_{\Bcal i}| \leq b_k, \wb^t_{\Bcal i} \geq 0}{\operatorname{argmin\hspace{0.7mm}}} \lVert \sum_{l \in \Scal^t_{\Bcal i}} \wb^t_{\Bcal il} \nabla_{\theta} L_T^{\Bcal_l}(\theta^t_i) - \nabla_{\theta} L_T^{\Bcal}(\theta^t_i) \rVert + \lambda \left \Vert \wb_{\Bcal i}^t \right \Vert ^2
\end{equation}

In the per-batch version, because the number of samples required for selection is $b_k$ is less than $k$, the number of greedy iterations required for data subset selection in OMP is reduced, resulting in a speedup of $B \times$. A critical trade-off in using larger batch sizes is that in order to get better speedups, we must also sacrifice data subset selection performance. Therefore, it is recommended to use smaller batch sizes for subset selection to get a optimal trade-off between speedups and performance. In our experiments on Image datasets, we use a batch size of $B=20$, and on text datasets, we use the batch size as a hyper-parameter with $B \in [16, 32, 64]$. Apart from per-batch selection, we use model warm-starting to get more informative data subsets. Further, in our experiments, we use a regularization coefficient of $\lambda=0$. We give more details on warm-starting below. 

\noindent \textbf{Warm-starting data selection: } We warm-start each configuration model by training on the entire training dataset for a few epochs similar to~\cite{killamsetty2021grad}. The warm-starting process enables the model to have informative loss gradients used for subset selection. To be more specific, the classifier model is trained on the entire training data for $T_w = \frac {\kappa T k}{N}$ epochs, where $k$ is the coreset size, $T$ is the total number of epochs, $\kappa$ is the fraction of warm start, and $N$ is the size of the training dataset. We use a $\kappa$ value of $0$ ({\em i.e.}, no warm start) for experiments using Hyperband as scheduling algorithm, and a $\kappa$ value of $0.35$ for experiments using ASHA.

\subsection{Component-3: Hyper-parameter Scheduling Algorithm}
Hyper-parameter scheduling algorithms improve the overall efficiency of the hyper-parameter tuning by terminating some of the poor configurations runs early. In our experiments, we consider Hyperband~\cite{hyperband}, and ASHA~\cite{asha}, which are extensions of the Sequential Halving algorithm (SHA)~\cite{sha} that uses aggressive early stopping to terminate poor configuration runs and allocates an increasingly exponential amount of resources to the better performing configurations. SHA starts with $n$ number of initial configurations, each assigned with a minimum resource amount $r$. The SHA algorithm uses a reduction factor $\eta$ to reduce the number of configurations in each round by selecting the top $\frac{1}{\eta}^{th}$ fraction of configurations while also increasing the resources allocated to these configurations by $\eta$ times each round. We discuss Hyperband and ASHA and the issues within SHA that each of them addresses in more detail in Appendix~\ref{app:hyperparamscheduler}.

Detailed pseudocode of the \model{} algorithm is provided in Appendix~\ref{app:pseudocode} due to space constraints in the main paper. We use the popular deep learning framework~\cite{pytorch} for implementation of \model{} framework, Ray-tune\cite{liaw2018tune} for hyper-parameter search and scheduling algorithms, and CORDS~\cite{Killamsetty_CORDS_COResets_and_2021} for subset selection strategies. 

\begin{figure}[!htbp]
\centering
\includegraphics[width = 14cm, height=0.6cm]{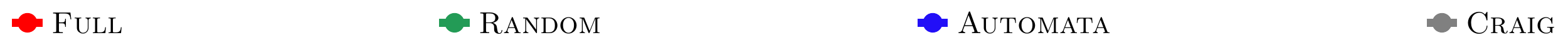}
\centering
\hspace{-0.6cm}
\begin{subfigure}[b]{0.24\textwidth}
\centering
\includegraphics[width=3.2cm, height=2.5cm]{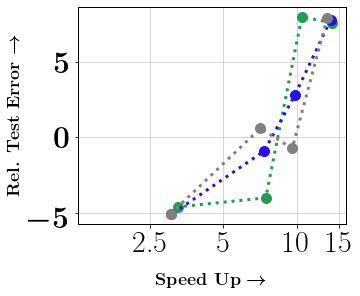}
\caption*{\scriptsize a) SST5(Random,HB)}
\phantomcaption
\label{fig:sst5_random_hb}
\end{subfigure}
\begin{subfigure}[b]{0.24\textwidth}
\centering
\includegraphics[width=3.2cm, height=2.5cm]{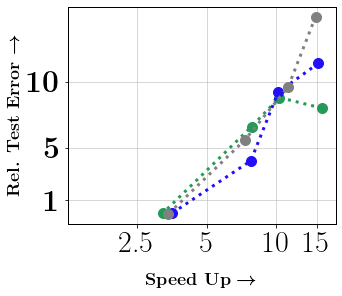}
\caption*{\scriptsize b) SST5(TPE,HB)}
\phantomcaption
\label{fig:sst5_tpe_hb}
\end{subfigure}
\begin{subfigure}[b]{0.24\textwidth}
\centering
\includegraphics[width=3.2cm, height=2.5cm]{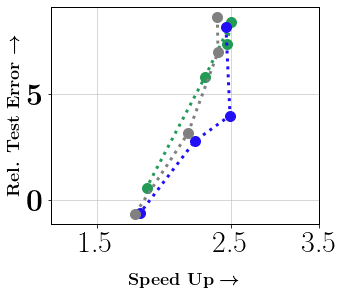}
\caption*{\scriptsize c) SST5(Random,ASHA)}
\phantomcaption
\label{fig:sst5_random_asha}
\end{subfigure}
\begin{subfigure}[b]{0.24\textwidth}
\centering
\includegraphics[width=3.2cm, height=2.5cm]{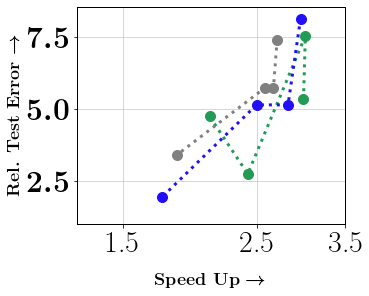}
\caption*{\scriptsize d) SST5(TPE,ASHA)}
\phantomcaption
\label{fig:sst5_tpe_asha}
\end{subfigure}
\hspace{-0.6cm}
\begin{subfigure}[b]{0.24\textwidth}
\centering
\includegraphics[width=3.2cm, height=2.5cm]{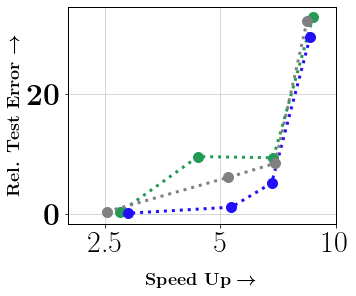}
\caption*{\scriptsize e) TREC6(Random,HB)}
\phantomcaption
\label{fig:trec6_random_hb}
\end{subfigure}
\begin{subfigure}[b]{0.24\textwidth}
\centering
\includegraphics[width=3.2cm, height=2.5cm]{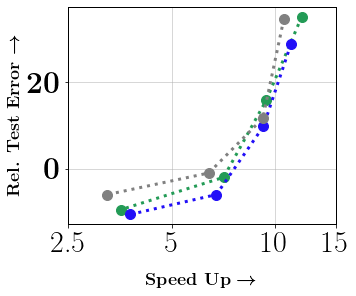}
\caption*{\scriptsize f) TREC6(TPE,HB)}
\phantomcaption
\label{fig:trec6_tpe_hb}
\end{subfigure}
\begin{subfigure}[b]{0.24\textwidth}
\centering
\includegraphics[width=3.2cm, height=2.5cm]{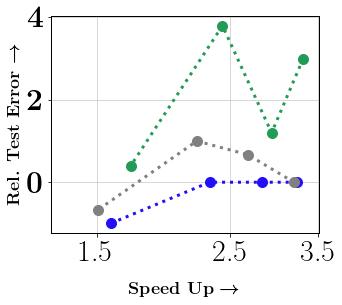}
\caption*{\scriptsize g) TREC6(Random,ASHA)}
\phantomcaption
\label{fig:trec6_random_asha}
\end{subfigure}
\begin{subfigure}[b]{0.24\textwidth}
\centering
\includegraphics[width=3.2cm, height=2.5cm]{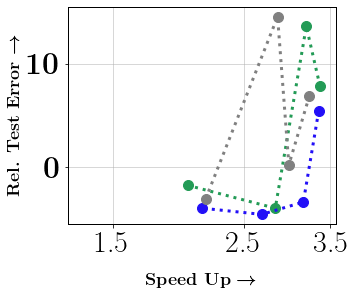}
\caption*{\scriptsize h) TREC6(TPE,ASHA)}
\phantomcaption
\label{fig:trec6_tpe_asha}
\end{subfigure}
\hspace{-0.6cm}
\begin{subfigure}[b]{0.24\textwidth}
\centering
\includegraphics[width=3.2cm, height=2.5cm]{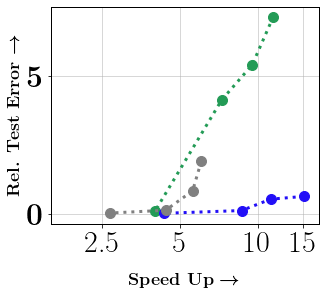}
\caption*{\scriptsize i) CIFAR10(Random,HB)}
\phantomcaption
\label{fig:cifar10_random_hb}
\end{subfigure}
\begin{subfigure}[b]{0.24\textwidth}
\centering
\includegraphics[width=3.2cm, height=2.5cm]{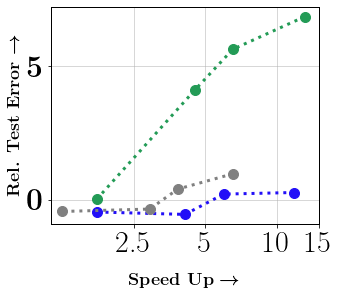}
\caption*{\scriptsize j) CIFAR10(TPE,HB)}
\phantomcaption
\label{fig:cifar10_tpe_hb}
\end{subfigure}
\begin{subfigure}[b]{0.24\textwidth}
\centering
\includegraphics[width=3.2cm, height=2.5cm]{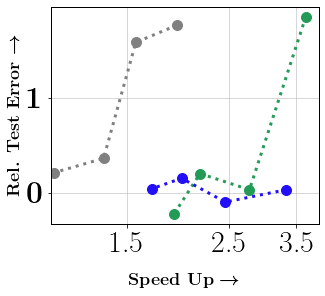}
\caption*{\scriptsize k) CIFAR10(Random,ASHA)}
\phantomcaption
\label{fig:cifar10_random_asha}
\end{subfigure}
\begin{subfigure}[b]{0.24\textwidth}
\centering
\includegraphics[width=3.2cm, height=2.5cm]{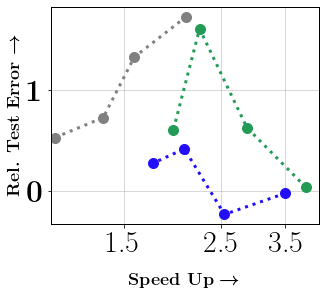}
\caption*{\scriptsize l) CIFAR10(TPE,ASHA)}
\phantomcaption
\label{fig:cifar10_tpe_asha}
\end{subfigure}
\hspace{-0.6cm}
\begin{subfigure}[b]{0.24\textwidth}
\centering
\includegraphics[width=3.2cm, height=2.5cm]{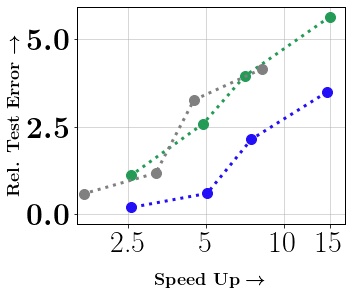}
\caption*{\scriptsize m) CIFAR100(Random,HB)}
\phantomcaption
\label{fig:cifar100_random_hb}
\end{subfigure}
\begin{subfigure}[b]{0.24\textwidth}
\centering
\includegraphics[width=3.2cm, height=2.5cm]{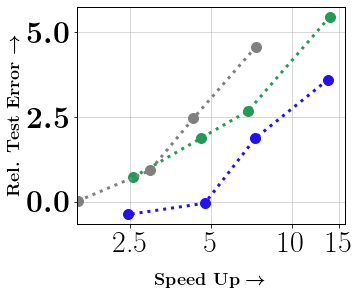}
\caption*{\scriptsize n) CIFAR100(TPE,HB)}
\phantomcaption
\label{fig:cifar100_tpe_hb}
\end{subfigure}
\begin{subfigure}[b]{0.24\textwidth}
\centering
\includegraphics[width=3.2cm, height=2.5cm]{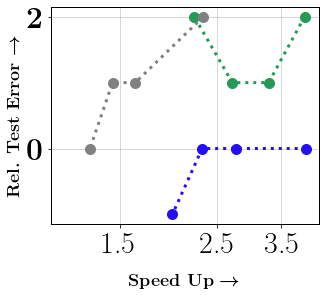}
\caption*{\scriptsize o) CIFAR100(Random,ASHA)}
\phantomcaption
\label{fig:cifar100_random_asha}
\end{subfigure}
\begin{subfigure}[b]{0.24\textwidth}
\centering
\includegraphics[width=3.2cm, height=2.5cm]{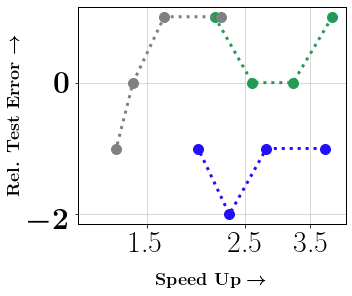}
\caption*{\scriptsize p) CIFAR100(TPE,ASHA)}
\phantomcaption
\label{fig:cifar100_tpe_asha}
\end{subfigure}
\hspace{-0.6cm}
\begin{subfigure}[b]{0.24\textwidth}
\centering
\includegraphics[width=3.2cm, height=2.5cm]{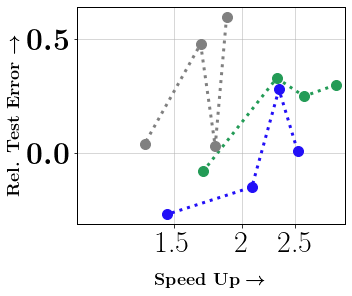}
\caption*{\scriptsize q) CONNECT-4(Random,HB)}
\phantomcaption
\label{fig:connect4_random_hb}
\end{subfigure}
\begin{subfigure}[b]{0.24\textwidth}
\centering
\includegraphics[width=3.2cm, height=2.5cm]{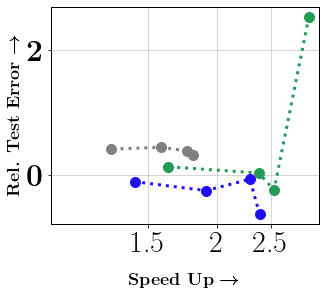}
\caption*{\scriptsize r) CONNECT-4(TPE,HB)}
\phantomcaption
\label{fig:connect4_tpe_hb}
\end{subfigure}
\begin{subfigure}[b]{0.24\textwidth}
\centering
\includegraphics[width=3.2cm, height=2.5cm]{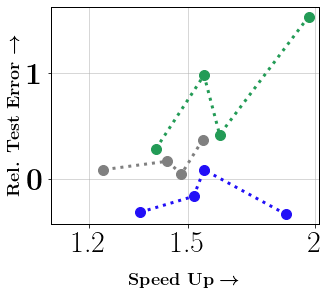}
\caption*{\scriptsize s) CONNECT-4(Rand,ASHA)}
\phantomcaption
\label{fig:connect4_random_asha}
\end{subfigure}
\begin{subfigure}[b]{0.24\textwidth}
\centering
\includegraphics[width=3.2cm, height=2.5cm]{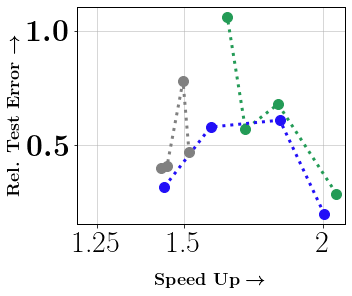}
\caption*{\scriptsize t) CONNECT-4(TPE,ASHA)}
\phantomcaption
\label{fig:connect4_tpe_asha}
\end{subfigure}
\caption{Comparison of  performance of \model\ with baselines(\textsc{Random, Craig, Full}) for Hyper-parameter tuning. In sub-figures (a-t), we present speedup {\em vs.} relative test error (in \%), compared to Full data tuning for different methods. On each scatter plot, smaller subsets appear on the right, and larger ones appear on the left. Results are shown for (a-d) SST5, (e-h) TREC6, (i-l) CIFAR10, (m-p) CIFAR100, and (q-t) CONNECT-4 datasets with different combinations of hyper-parameter search and scheduling algorithms. \emph{The scatter plots show that \model{} achieves the best speedup-accuracy tradeoff in almost every case  \textbf{(bottom-right corner of each plot indicates the best speedup-accuracy tradeoff region})}. }
\label{fig:main_results}
\end{figure}

\section{Experiments} \label{sec:experiments}
In this section, we present the effectiveness and the efficiency of \model\ framework for hyper-parameter tuning by evaluating \model\ on datasets spanning text, image, and tabular domains. Further, to assess \model{}'s effectiveness across the spectrum of existing hyper-parameter search and scheduling algorithms, we conduct experiments using combinations of different search and scheduling algorithms. As discussed earlier, we employ Random Search~\cite{randomsearch}, TPE~\cite{tpe} as representative hyper-parameter search algorithms, and Hyperband~\cite{hyperband}, ASHA~\cite{asha} as representative hyper-parameter scheduling algorithms. However, we believe the takeaways would remain the same even with other approaches. We repeat each experiment five times on the text and tabular datasets, thrice on the image datasets, and  report the mean accuracy and speedups in the plots. Below, we provide further details on datasets, baselines, models, and the hyper-parameter search space used for experiments.

\subsection{Baselines} 
Our experiments aim to demonstrate the consistency and efficiency of \model\, more specifically, the effectiveness of \model's gradient-based subset selection (GSS) for hyper-parameter tuning. As baselines, we replace the GSS subset selection strategy in \model\ with different subset selection strategies, namely \textsc{Random} (randomly sample a same sized subset as \model\ from the training data), \textsc{Craig}~\cite{mirzasoleiman2020coresets} (a gradient-based subset selection proposed for efficient supervised learning), and \textsc{Full} (using the entire training data for model training during configuration evaluation). For ease of notation, we refer to baselines by the names of corresponding subset selection strategies. Note that by \textsc{Craig} baseline, we mean the faster per-batch version of CRAIG~\cite{mirzasoleiman2020coresets} for subset selection shown~\cite{killamsetty2021grad} to be more efficient than the original. In addition, for all methods, we do not use any warm-start for experiments with Hyperband and use a warm start of $\kappa=0.35$ for experiments with ASHA. We give more details on the reason for using warm-start with ASHA and no warm-start with Hyperband in Appendix~\ref{app:warmstart}. We perform experiments with different subset size fractions of 1\%, 5\%, 10\%, and 30\%. In our experiments, we compare our approach's accuracy and efficiency (time/energy) with Full training, Per Batch CRAIG selection, and Random selection.

\subsection{Datasets, Model Architecture, and Experimental Setup} 
To demonstrate the effectiveness of \model\ for hyper-parameter tuning, we performed experiments on datasets spanning text, image and tabular domains. Text datasets include SST2~\cite{sst}, SST5~\cite{sst}, glue-SST2~\cite{gluesst2}, and TREC6~\cite{trec1, trec2}.
Image datasets include CIFAR10~\cite{Krizhevsky09learningmultiple}, CIFAR100~\cite{Krizhevsky09learningmultiple}, and  Street View House Numbers (SVHN)~\cite{Netzer2011}.
Tabular datasets include DNA, SATIMAGE, LETTER, and CONNECT-4 from \textbf{LIBSVM} (a library for Support Vector Machines (SVMs))~\cite{CC01a}. We give more details on dataset sizes and splits in Appendix~\ref{app:datasets}. For the Text datasets, we use the LSTM model (from PyTorch) with trainable GloVe~\cite{pennington-etal-2014-glove}
embeddings of 300 dimension as input. For Image datasets, we use the ResNet18~\cite{He2016DeepRL} and ResNet50~\cite{He2016DeepRL} models. For Tabular datasets, we use a multi-layer perceptron with 2 hidden layers.
Once the best hyper-parameter configuration is found, we perform one more \textit{final training} of the model using the best configuration on the entire dataset and report the achieved test accuracy. We use \textit{final training} for all methods except \textsc{Full} since the models trained on small data subsets (especially with small subset fractions of 1\%, 5\%) during tuning do not achieve high test accuracies. We also include the final training times while calculating the tuning times for a more fair comparison\footnote{Note that with a 30\% subset, final training is not required as the models trained with 30\% subsets achieve similar accuracy to full data training. However, for the sake of consistency, we use \textit{final training} with 30\% subsets as well.} For text datasets, we train the LSTM model for 20 epochs while choosing subsets (except for \textsc{Full}) every 5 epochs. The hyper-parameter space includes learning rate, hidden size \& number of layers of LSTM, batch size of training. Some experiments (with TPE as the search algorithm) use 27 configurations in the hyper-parameter space, while others use 54. More details on hyper-parameter search space for text datasets are given in Appendix~\ref{app:textexperimentaldetails}. For image datasets, we train the ResNet~\cite{He2016DeepRL} model for 300 epochs while choosing subsets (except for \textsc{Full}) every 20 epochs. We use a Stochastic Gradient Descent (SGD) optimizer with momentum set to 0.9 and weight decay factor set to 0.0005. The hyper-parameter search space consists of a choice between the Momentum method and Nesterov Accelerated Gradient method, choice of learning rate scheduler and their corresponding parameters, and four different group-wise learning rates. We use 27 configurations in the hyper-parameter space for Image datasets. More details on hyper-parameter search space for image datasets are given in Appendix~\ref{app:imageexperimentaldetails}. For tabular datasets, we train a multi-layer perceptron with 2 hidden layers for 200 epochs while choosing subsets every 10 epochs. The hyper-parameter search space consists of a choice between the SGD optimizer or Adam optimizer, choice of learning rate, choice of learning rate scheduler, the sizes of the two hidden layers and batch size for training. We use 27 configurations in the hyper-parameter space for Tabular datasets. More details on hyper-parameter search space for tabular datasets are provided in Appendix~\ref{app:tableexperimentaldetails}.

\subsection{Hyper-parameter Tuning Results} 
Results comparing the accuracy vs. efficiency tradeoff of different subset selection strategies for hyper-parameter tuning are shown in \figref{fig:main_results}. Performance is compared for different sizes of subsets of training data: 1\%, 5\%, 10\%, and 30\% along with four possible combinations of search algorithm (Random or TPE) and scheduling algorithm (ASHA or Hyperband). \noindent \textbf{Text datasets results: } Sub-figures(\ref{fig:sst5_random_hb}, \ref{fig:sst5_tpe_hb}, \ref{fig:sst5_random_asha}, \ref{fig:sst5_tpe_asha}) show the plots of relative test error {\em vs.} speed ups, both {\em w.r.t} full data tuning for SST5 dataset with different combinations of search and scheduling methods. 
Similarly, in sub-figures(\ref{fig:trec6_random_hb}, \ref{fig:trec6_tpe_hb}, \ref{fig:trec6_random_asha}, \ref{fig:trec6_tpe_asha}) we present  the plots of relative test error {\em vs.} speed ups for TREC6 dataset. From the results, we observe that \model\ achieves best speed up {\em vs.} accuracy tradeoff and consistently gives better performance even with small subset sizes unlike other baselines like \textsc{Random, Craig}. In particular, \model\ achieves a speedup of 9.8$\times$ and 7.35$\times$ with a performance loss of 2.8\% and a performance gain of 0.9\% respectively on the SST5 dataset with TPE and Hyperband. 
Additionally, \model\ achieves a speedup of around 3.15$\times$, 2.68$\times$ with a performance gain of 3.4\%, 4.6\% respectively for the TREC6 dataset with TPE and ASHA. \noindent \textbf{Image datasets results: } Sub-figures(\ref{fig:cifar10_random_hb}, \ref{fig:cifar10_tpe_hb}, \ref{fig:cifar10_random_asha}, \ref{fig:cifar10_tpe_asha}) show the plots of relative test error {\em vs.} speed ups, both {\em w.r.t} full data tuning for CIFAR10 dataset with different combinations of search and scheduling methods. Similarly, sub-figures (\ref{fig:cifar100_random_hb}, \ref{fig:cifar100_tpe_hb}, \ref{fig:cifar100_random_asha}, \ref{fig:cifar100_tpe_asha}) show the plots of relative test error {\em vs.} speed ups on CIFAR100. The results show that \model\ achieves the best speed up {\em vs.} accuracy tradeoff consistently compared to other baselines. More specifically, \model\ achieves a speedup of around 15$\times$, 8.7$\times$ with a performance loss of $0.65\%$, $0.14\%$ respectively on the CIFAR10 dataset with Random and Hyperband. Further, \model\ achieves a speedup of around 3.7$\times$, 2.3$\times$ with a performance gain of $1\%$, $2\%$ for CIFAR100 dataset with TPE and ASHA. \noindent \textbf{Tabular datasets results: } Sub-figures(\ref{fig:connect4_random_hb}, \ref{fig:connect4_tpe_hb}, \ref{fig:connect4_random_asha}, \ref{fig:connect4_tpe_asha}) show the plots of relative test error {\em vs.} speed ups for the CONNECT-4 dataset. \model\ consistently achieved better speedup {\em vs.} accuracy tradeoff compared to other baselines on CONNECT-4 as well. Owing to space constraints, we provide additional results showing the accuracy {\em vs.} efficiency tradeoff on additional text, image, and tabular datasets in the Appendix~\ref{app:experimentalresults}. It is important to note that \model\ obtains better speedups when used for hyper-parameter tuning on larger datasets and larger models (in terms of parameters). 
Apart from the speedups achieved by \model, we show in Appendix~\ref{app:co2emissions} that it also achieves similar reductions of energy consumption and CO2 emissions, thereby making it more environmentally friendly. 

\section{Conclusion, Limitations, and Broader Impact} \label{sec:conclusion}
We introduce \model{}, an efficient hyper-parameter tuning framework
that uses intelligent subset selection for model training for faster configuration evaluations. Further, we perform extensive experiments showing the effectiveness of \model\ for Hyper-parameter tuning. In particular, it achieves speedups of around $10\times$ - $15\times$ using Hyperband as scheduler and speedups of around $3 \times$ even with a more efficient ASHA scheduler. \model\ significantly decreases CO2 emissions by making hyper-parameter tuning fast and energy-efficient, in turn reducing environmental impact of such hyper-parameter tuning on society at large. We hope that the \model\ framework will popularize the trend of using subset selection for hyper-parameter tuning and encourage further research on  efficient subset selection approaches for faster hyper-parameter tuning, helping us move closer to the goal of Green AI~\cite{Schwartz2020GreenA}. One of the limitations of \model\ is that in scenarios in which no performance loss is desired, we do not know the minimum subset size to improve speed and, therefore, rely on larger subset sizes such as 10\%, 30\%. In the future, we consider adaptively changing subset sizes based on model performance for each configuration to remove the dependency on subset size.

\bibliography{main}

\appendix
\newpage

\begin{center}
    \Huge{Supplementary Material}
\end{center}
\section{Code} \label{app:code}
The code of \model{} is available at the following link: \url{https://github.com/decile-team/cords}.

\section{Licenses} \label{app:licenses}
We release the code repository of \model{} with MIT license, and it is available for everybody to use freely. We use the popular deep learning framework~\cite{pytorch} for implementation of \model{} framework, Ray-tune\cite{liaw2018tune} for hyper-parameter search and scheduling algorithms, and CORDS~\cite{Killamsetty_CORDS_COResets_and_2021} for subset selection strategies. As far as the datasets are considered, we use SST2~\cite{sst}, SST5~\cite{sst}, glue-SST2~\cite{gluesst2}, TREC6~\cite{trec1, trec2}, CIFAR10~\cite{Krizhevsky09learningmultiple}, SVHN~\cite{Netzer2011}, CIFAR100~\cite{Krizhevsky09learningmultiple}, and DNA, SATIMAGE, LETTER, CONNECT-4 from \textbf{LIBSVM} (a library for Support Vector Machines (SVMs))~\cite{CC01a} datasets. CIFAR10, CIFAR100 datasets are released with an MIT license. SVHN dataset is released with a CC0:Public Domain license. Furthermore, all the datasets used in this work are publicly available. In addition, the datasets used do not contain any personally identifiable information.

\section{\model{} Algorithm Pseudocode} \label{app:pseudocode}
We give the pseudo code of \model{} algorithm in Algorithm~\ref{alg:mainalg}. 
\begin{algorithm}[H]
\SetCustomAlgoRuledWidth{0.45\textwidth}
\LinesNotNumbered
\DontPrintSemicolon
\KwIn{Hyper-parameter scheduler Algorithm: $\operatorname{scheduler\hspace{0.7mm}}$, Hyper-parameter search Algorithm: $\operatorname{search\hspace{0.7mm}}$,
       No. of configuration evaluations: $n$, Hyper-parameter search space: $\Hcal$, Training dataset: $\Dcal$, Validation dataset: $\Vcal$,
       Total no of epochs: $T$, Epoch interval for subset selection: $R$, Size of the coreset: $k$, 
       Reg. Coefficient: $\lambda$, Learning rates: $\{\alpha_t\}_{t=0}^{t=T-1}$, Tolerance: $\epsilon$}
 \SetKwBlock{Begin}{function}{end function}{
    Generate $n$ configurations by calling the search algorithm $H = \{h_1, h_2, \cdots, h_n\} = \operatorname{search\hspace{0.7mm}}(\Hcal, n)$ \;
    Randomly initialize each configuration model parameters $h_1.\theta = h_2.\theta = \cdots = h_n.\theta = \theta$\;
    Set $h_1.t= h_2.t = \cdots = h_n.t = 0$; \;
    Set $h_1.eval= h_2.eval = \cdots = h_n.eval = 0$; \;
    Assign the initial resources(i.e., in our case training epochs) using the scheduler for all initialized configurations 
    $\{h_i.r\}_{i=1}^{i=n} = \operatorname{scheduler\hspace{0.7mm}}(H, T)$ \;
    \Repeat{until $h_1.r == 0 \And h_2.r == 0 \And \cdots h_n.r == 0$}
    {
        \textcolor{gray}{***Evaluate all remaining configurations*** } \;
        \For{each configuration numbered $i$ in $H$}
        {   
        \textcolor{gray}{***Train configuration $h_i$ using informative data subsets for $h_i.r$ epochs and evaluate on validation set*** } \;
         $h_i.eval, h_i.theta$ = $\operatorname{subset-config-evaluation\hspace{0.7mm}}(\Dcal, \Vcal, h_i.theta, h_i.r, R, k, \lambda, \{\alpha_t\}_{t=0}^{t=h_i.r}, \epsilon)$\;
         $h_i.t = h_i.t + h_i.r$\;
        }
        \textcolor{gray}{***Assign resources again based on evaluation performance*** } \;
        $\{h_i.r\}_{i=1}^{i=n} = \operatorname{scheduler\hspace{0.7mm}}(H, T)$ \;
    }
    \textcolor{gray}{***Get the best performing hyper-parameters based on final configuration evaluations***}
    $final\-config = \underset{h_i.config}{\argmax}[h_i.eval]_{i=1}^{n}$
    \textcolor{gray}{***Perform final training using the best hyper-parameter configurations***}
    $\theta_{final} = \operatorname{final\-train\hspace{0.7mm}}(\theta, \Dcal, final\-config, T)$
 }
 \Return{$\theta_{final}$}
\caption{\model{} Algorithm}\label{alg:mainalg}
\end{algorithm}

\begin{algorithm}[H]
\SetCustomAlgoRuledWidth{0.45\textwidth}
\LinesNotNumbered
 \DontPrintSemicolon
 \KwIn{Training dataset: $\Dcal$, Validation dataset: $\Vcal$, Initial model parameters: $\theta_0$, 
      Total no of epochs: $T$, Epoch interval for subset selection: $R$, Size of the coreset: $k$,
      Reg. Coefficient: $\lambda$, Learning rates: $\{\alpha_t\}_{t=0}^{t=T-1}$, Tolerance: $\epsilon$}
 \SetKwBlock{Begin}{function}{end function}{
    Set $t=0$; Randomly initialize coreset $\Scal_{0} \subseteq \Dcal: |\Scal_{0}| = k$; \;
    \Repeat{until $t \ge T$}
    {
        \If {$(t\%R == 0) \land (t > 0)$}{
        $\Scal_{t}$ = OMP($\Dcal, \theta_{t}, \lambda, \alpha_{t}, k, \epsilon$)\;} 
        \Else { $\Scal_{t} = \Scal_{t-1}$\;}
        Compute batches $\Dcal_b = ((x_b, y_b); b \in (1 \cdots B))$ from $\Dcal$ \;
        Compute batches $\Scal_{tb} = ((x_b); b \in (1 \cdots B))$ from $\Scal$ \;
        \textcolor{gray}{*** Mini-batch SGD *** } \;
        Set $\theta_{t0} = \theta_{t}$ \;
        \For{$b = 1$ to $B$}
        {   
        Compute mask $\mb_{t}$ on $\Scal_{tb}$ from current model parameters $\theta_{t(b-1)}$\;
        $\theta_{tb} = \theta_{t(b-1)} - \alpha_t \nabla_{\theta}L_S(\Dcal_b, \theta_t) - \alpha_t \lambda_t \underset{j \in \Scal_{tb}}{\sum} \mb_{jt} \nabla_{\theta}l_u(x_j, \theta_t(b-1))$ \;
        }
        Set $\theta_{t+1} = \theta_{tB}$ \;
        $t = t+1$\;
    }
    \textcolor{gray}{*** Evaluate trained model on validation set *** } \;
    $eval = \operatorname{evaluate\hspace{0.7mm}}(\theta_T, \Vcal)$ \;
 }
 \Return{$eval, \theta_{T}$}
\caption{subset-config-evaluation}\label{alg:subset_evaluation}
\end{algorithm}

\begin{algorithm}[H]
\SetCustomAlgoRuledWidth{0.45\textwidth}
\LinesNotNumbered
\DontPrintSemicolon
\KwIn{Training loss $L_T$, current parameters: $\theta$, regularization coefficient: $\lambda$, subset size: $k$, tolerance: $\epsilon$}
\SetKwBlock{Begin}{function}{end function}{
    Initialize $\Scal = \emptyset$ \;
    ${r}\leftarrow \nabla_w (\Vert \sum_{l \in \Scal} \wb \nabla_{\theta}L_T^l(\theta) -  \nabla_{\theta}L_T(\theta)\Vert  + \lambda \left \Vert \wb \right \Vert ^2|)_{{\wb}={0}}$\;
    \Repeat{until $|\Scal| \leq k$ and $ \Vert \sum_{l \in \Scal} \wb \nabla_{\theta}L_T^l(\theta) -  \nabla_{\theta}L_T(\theta)\Vert  + \lambda \left \Vert \wb \right \Vert ^2 \geq \epsilon$}
    {
    $e = \text{argmax}_j |r_j|$ \;
    $\Scal\leftarrow \Scal \cup \{e\}$ \;
    ${\wb}\leftarrow \text{argmin}_{{\wb}}(\Vert \sum_{l \in \Scal} \wb \nabla_{\theta}L_T^l(\theta) -  \nabla_{\theta}L_T(\theta)\Vert  + \lambda \left \Vert \wb \right \Vert^2)$ \;
    ${r}\leftarrow \nabla_{\wb} (\Vert \sum_{l \in \Scal} \wb \nabla_{\theta}L_T^l(\theta) -  \nabla_{\theta}L_T(\theta)\Vert  + \lambda \left \Vert \wb \right \Vert ^2|)$ \;
    }
}
\Return{$\Scal, \wb$}
\caption{OMP}\label{alg:omp}
\end{algorithm}

\section{More details on Hyper-parameter Search Algorithms} \label{app:hyperparamsearch}
We give a brief overview of few representative hyper-parameter search algorithms, such as TPE~\cite{tpe} and Random Search~\cite{randomsearch} which we used in our experiments. As discussed earlier, given a hyper-parameter search space, hyper-parameter search algorithms provide a set of configurations that need to be evaluated. A naive way of performing the hyper-parameter search is Grid Search, which defines the search space as a grid and exhaustively evaluates each grid configuration. However, Grid Search is a time-consuming process, meaning that thousands to millions of configurations would need to be evaluated if the hyper-parameter space is large. In order to find optimal hyper-parameter settings quickly, Bayesian optimization-based hyper-parameter search algorithms have been developed. To investigate the effectiveness of \model\ across the spectrum of search algorithms, we used the Random Search method and the Bayesian optimization-based TPE method (described below) as representative hyper-parameter search algorithms.

\subsection{Random Search} In random search~\cite{randomsearch}, hyper-parameter configurations are selected at random and evaluated to discover the optimal configuration among those chosen. As well as being more efficient than a grid search since it does not evaluate all possible configurations exhaustively, random search also reduces overfitting~\cite{JMLR:v13:bergstra12a}. 


\subsection{Tree Parzen Structured Estimator (TPE)} TPE~\cite{tpe} is a sequential model-based optimization (SMBO) approach that sequentially constructs a probability model to approximate the performance of hyper-parameters based on historical configuration evaluations and then subsequently uses the model to select new configurations. TPE models the likelihood function $P(D|f)$ and the prior over the function space $P(f)$ using the kernel density estimation. TPE algorithm sorts the collected observations by the function evaluation value, typically validation set performance, and divides them into two groups based on some quantile. The first group $x_1$ contains best-performing observations, and the second group $x_2$ contains all other observations. Then TPE models two different densities $i(x_1)$ and $g(x_2)$ based on the observations from the respective groups using kernel density estimation. Finally, TPE selects the subset observations that need to be evaluated by sampling from the distribution that models the maximum expected improvement, i.e., $\EE[i(x)/g(x)]$.

\section{More details on Hyper-parameter Scheduling Algorithms} \label{app:hyperparamscheduler}
We give a brief overview of some representative hyper-parameter scheduling algorithms, such as HyperBand~\cite{hyperband} and ASHA~\cite{asha} which we used in our experiments. As discussed earlier, hyper-parameter scheduling algorithms improve the overall efficiency of the hyper-parameter tuning by terminating some of the poor configurations runs early. In our experiments, we consider Hyperband,  and ASHA, which are extensions of the Sequential Halving algorithm (SHA)~\cite{sha} that uses aggressive early stopping to terminate poor configuration runs and allocates an increasingly exponential amount of resources to the better performing configurations. SHA starts with $n$ number of initial configurations, each assigned with a minimum resource amount $r$. The SHA algorithm uses a reduction factor $\eta$ to reduce the number of configurations each round by selecting the top $\frac{1}{\eta}^{th}$ fraction of configurations while also increasing the resources allocated to these configurations by $\eta$ times each round. Following, we will discuss Hyperband and ASHA and the issues within SHA that each of them addresses.

\subsection{HyperBand} One of the issues with SHA is that its performance largely depends on the initial number $n$ of configurations. Hyperband~\cite{hyperband} addresses this issue by performing a grid search over various feasible values of $n$. Further, each value of $n$ is associated with a minimum resource $r$ allocated to all configurations before some are terminated; larger values of $n$ are assigned smaller $r$ and hence more aggressive early-stopping. On the whole, in Hyperband~\cite{hyperband} for different values of $n$ and $r$, the SHA algorithm is run until completion.

\subsection{ASHA: } One of the other issues with SHA is that the algorithm is sequential and has to wait for all the processes (assigned with an equal amount of resources) at a particular bracket to be completed before choosing the configurations to be selected for subsequent runs. Hence, due to the sequential nature of SHA, some GPU/CPU resources (with no processes running) cannot be effectively utilized in the distributed training setting, thereby taking more time for tuning. By contrast, ASHA~\cite{asha} is an asynchronous variant of SHA and addresses the sequential issue of SHA by promoting a configuration to the next rung as long as there are GPU or CPU resources available. If no resources appear to be promotable, it randomly adds a new configuration to the base rung.

\section{More Experimental Details and Additional Results}
We performed experiments on a mix of RTX 1080, RTX 2080, and V100 GPU servers containing 2-8 GPUs. To be fair in timing computation, we ran \model\ and all other baselines for a particular setting on the same GPU server.
\subsection{Additional Datasets Details}\label{app:datasets}
\subsubsection{Text Datasets}
We performed experiments on SST2~\cite{sst}, SST5~\cite{sst}, glue-SST2~\cite{gluesst2}, and TREC6~\cite{trec1, trec2} text datasets. SST2~\cite{sst} and glue-SST2~\cite{gluesst2} dataset classify the sentiment of the sentence (movie reviews) as negative or positive. Whereas SST5~\cite{sst} classify sentiment of sentence as negative, somewhat negative, neutral, somewhat positive or positive. TREC6~\cite{trec1, trec2} is a dataset for question classification consisting of open-domain, fact-based questions divided into broad semantic categories(ABBR - Abbreviation, DESC - Description and abstract concepts, ENTY - Entities, HUM - Human beings, LOC - Locations, NYM - Numeric values). The train, text and validation splits for SST2~\cite{sst} and SST5~\cite{sst} are used from the source itself while the validation data for TREC6~\cite{trec1, trec2} is obtained using 10\% of the train data. The train and validation data for glue-SST2~\cite{gluesst2} is used from source itself. In Table~\ref{tab:textdatasplits}, we summarize the number classes, and number of instances in each split in the text datasets.
\begin{table}[!htbp]
    \centering
    \begin{tabular}{|l|l|l|l|l|}
    \hline
        Dataset & \#Classes & \#Train & \#Validation & \#Test \\ \hline
        SST2 & 2 & 8544 & 1101 & 2210 \\ \hline
        SST5 & 5 & 8544 & 1101 & 2210 \\ \hline
        glue-SST2 & 2 & 63982 & 872 & 3367 \\ \hline
        TREC6 & 6 & 4907 & 545 & 500 \\ \hline
    \end{tabular}
    \caption{Number of classes, Number of instances in Train, Validation and Test split in Text datasets\label{tab:textdatasplits}}
\end{table}

\subsubsection{Vision Datasets}
We performed experiments on CIFAR10~\cite{Krizhevsky09learningmultiple}, CIFAR100~\cite{Krizhevsky09learningmultiple}, and SVHN~\cite{Netzer2011} vision datasets. The CIFAR-10~\cite{Krizhevsky09learningmultiple} dataset contains 60,000 colored images of size 32$\times$32 divided into ten classes, each with 6000 images. CIFAR100~\cite{Krizhevsky09learningmultiple} is also similar but that it has 600 images per class and 100 classes. Both CIFAR10~\cite{Krizhevsky09learningmultiple} and CIFAR100~\cite{Krizhevsky09learningmultiple} have 50,000 training samples and 10,000 test samples distributed equally across all classes. SVHN~\cite{Netzer2011} is obtained from house numbers in Google Street View images and has 10 classes, one for each digit. The colored images of size 32$\times$32 are centered around a single digit with some distracting characters on the side. SVHN~\cite{Netzer2011}  has 73,257 training digits, 26,032 testing digits. For all 3 datasets, $10\%$ of the training data is used for validation. In Table~\ref{tab:imagedatasplits}, we summarize the number classes, and number of instances in each split in the image datasets.

\begin{table}[!htbp]
    \centering
    \begin{tabular}{|l|l|l|l|l|}
    \hline
        Dataset & \#Classes & \#Train & \#Validation & \#Test \\ \hline
        CIFAR10 & 10 & 45000 & 5000 & 10000 \\ \hline
        CIFAR100 & 100 & 45000 & 5000 & 10000 \\ \hline
        SVHN &  10 & 65932 & 7325 & 26032\\ \hline
    \end{tabular}
    \caption{Number of classes, Number of instances in Train, Validation and Test split in Image datasets\label{tab:imagedatasplits}}
\end{table}

\subsubsection{Tabular Datasets}
We performed experiments on the following tabular datasets \textbf{dna, letter, connect-4, and  satimage} from \textbf{LIBSVM} (a library for Support Vector Machines (SVMs))~\cite{CC01a}. 

\begin{table}[!htbp]
  \centering
  \begin{tabular}{|c|c|c|c|c|c|} 
 \hline 
 \textbf{Name} & \textbf{\#Classes} & \textbf{\#Train} & \textbf{\#Validation} & \textbf{\#Test} & \textbf{\#Features} \\ [0.5ex] 
 \hline
 dna & 3 & 1,400 & 600 & 1,186 & 180\\ 
 \hline
 satimage & 6 & 3,104 & 1,331& 2,000 & 36\\ 
 \hline
 letter & 26 & 10,500 & 4,500 & 5,000 & 16\\ 
 \hline
 \text{connect\_4} & 3 & 67,557 & - & - & 126 \\ 
 \hline
 \end{tabular}
 \caption{Number of classes, Number of instances in Train, Validation and Test split in Tabular datasets\label{tab:tabulardatasplits}}
\end{table}

A brief description of the tabular datasets can be found in Table~\ref{tab:tabulardatasplits}. For datasets without explicit validation and test datasets, 10\% and 20\% samples from the training set are used as validation and test datasets, respectively. 

\subsection{Additional Experimental Details}\label{app:experimentaldetails}
For tuning with \textsc{Full} datasets, the entire dataset is used to train the model during hyper-parameter tuning. But when the \model\ (or \textsc{Craig}) is used, only a fraction of the dataset is used to train various models during tuning. Similar is the case with Random subset selection approach but the subsets are chosen at \textsc{Random}. Note that subset selection techniques used are adaptive in nature, which mean that they chose subset every few epochs for the model to train on for coming few epochs.
\subsubsection{Details of Text Experiments}\label{app:textexperimentaldetails}
The hyper-parameter space for experiments on text datasets include learning rate, hidden size \& number of layers of LSTM and batch size of training. Some experiments (with TPE search algorithm) where the best configuration among 27 configurations are found, the hyper-parameter space is learning rate: [0.001,0.1], LSTM hidden size: \{64,128,256\}, batch size: \{16,32,64\}. While the rest of the experiments where the best configuration among 54 configurations are found, the hyper-parameter space is learning rate: [0.001,0.1], LSTM hidden size: \{64,128,256\}, number of layers in LSTM: \{1, 2\}, batch size: \{16,32,64\}.

\subsubsection{Details of Image Experiments}\label{app:imageexperimentaldetails}
The hyper-parameter search space for tuning experiments on image datasets include a choice between Momentum method and Nesterov Accelerated Gradient method, choice of learning rate scheduler and their corresponding parameters, and four different group-wise learning rates, $lr_1$ for layers of the first group, $lr_2$ for layers of intermediate groups, $lr_3$ for layers of the last group of ResNet model, and $lr_4$ for the final fully connected layer. For learning rate scheduler, we change the learning rates during training using either a cosine annealing schedule or decay it linearly by $\gamma$ after every 20 epochs. Best configuration for most experiments is selected from 27 configurations where the hyper-parameter space is $lr_1$: [0.001, 0.01], $lr_2$: [0.001, 0.01], $lr_3$: [0.001, 0.01], $lr_4$: [0.001, 0.01], Nesterov: \{True, False\}, learning rate scheduler: \{Cosine Annealing, Linear Decay\}, $\gamma$: [0.05, 0.5].

\subsubsection{Details of Tabular Experiments}\label{app:tableexperimentaldetails}
The hyper-parameter search space consists of a choice between the Stochastic Gradient Descent(SGD) optimizer or Adam optimizer, choice of learning rate $lr$, choice of learning rate scheduler, the sizes of the two hidden layers $h_1$ and $h_2$ and batch size for training. For learning rate scheduler, we either don't use a learning rate scheduler or change the learning rates during training using a cosine annealing schedule or decay it linearly by 0.05 after every 20 epochs. Best configuration for most experiments is selected from 27 configurations where the hyper-parameter space is $lr$: [0.001, 0.01], Optimizer: \{Adam, SGD\}, learning rate scheduler: \{None, Cosine Annealing, Linear Decay\}, $h_1$: \{150, 200, 250, 300\},  $h_2$: \{150, 200, 250, 300\} and batch size: \{16,32,64\}.

\subsection{Use of Warm-start for subset selection} \label{app:warmstart}
We use warm-starting with ASHA as a scheduler because the initial bracket occurs early (i.e., at $t=1$) with ASHA; this implies that some of the initial configuration evaluations are discarded made just after training for one epoch. The training of such configurations on small data subsets may not be sufficient to make a sound decision about better-performing configurations in these scenarios. As a solution, we use warm-starting with ASHA so that all configurations are trained on the entire data for an initial few epochs. With Hyperband, the brackets do not occur very early during training, so no warm-up is necessary. 

\subsection{More Hyper-parameter Tuning Results} \label{app:experimentalresults}
\begin{figure}[t]
\centering
\includegraphics[width = 14cm, height=0.6cm]{figs/legend.pdf}
\centering
\hspace{-0.6cm}
\begin{subfigure}[b]{0.24\textwidth}
\centering
\includegraphics[width=3.2cm, height=2.5cm]{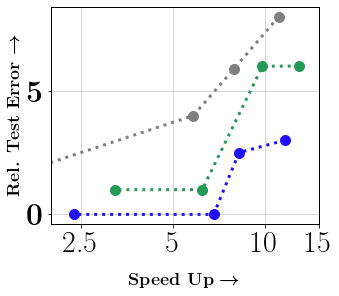}
\caption*{$\underbracket[1pt][1mm]{\hspace{3.6cm}}_{\substack{\vspace{-4.0mm}\\
\colorbox{white}{(a) \scriptsize SST2(Random,HB)}}}$}
\phantomcaption
\label{appfig:sst2_random_hb}
\end{subfigure}
\begin{subfigure}[b]{0.24\textwidth}
\centering
\includegraphics[width=3.2cm, height=2.5cm]{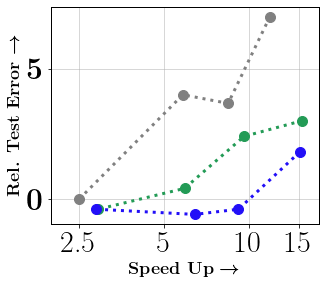}
\caption*{$\underbracket[1pt][1.0mm]{\hspace{3.6cm}}_{\substack{\vspace{-4.0mm}\\
\colorbox{white}{(b) \scriptsize SST2(TPE,HB)}}}$}
\phantomcaption
\label{appfig:sst2_tpe_hb}
\end{subfigure}
\begin{subfigure}[b]{0.24\textwidth}
\centering
\includegraphics[width=3.2cm, height=2.5cm]{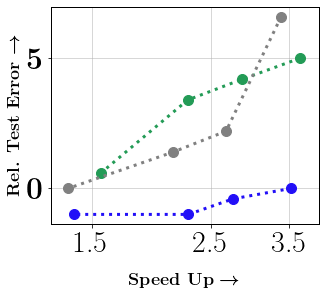}
\caption*{$\underbracket[1pt][1.0mm]{\hspace{3.6cm}}_{\substack{\vspace{-4.0mm}\\
\colorbox{white}{(c) \scriptsize SST2(Random,ASHA)}}}$}
\phantomcaption
\label{appfig:sst2_random_asha}
\end{subfigure}
\begin{subfigure}[b]{0.24\textwidth}
\centering
\includegraphics[width=3.2cm, height=2.5cm]{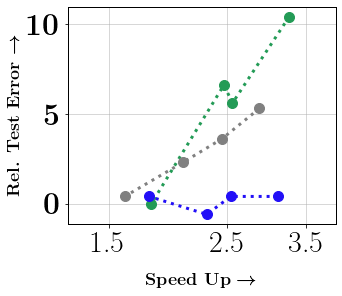}
\caption*{$\underbracket[1pt][1.0mm]{\hspace{3.6cm}}_{\substack{\vspace{-4.0mm}\\
\colorbox{white}{(d) \scriptsize SST2(TPE,ASHA)}}}$}
\phantomcaption
\label{appfig:sst2_tpe_asha}
\end{subfigure}
\begin{subfigure}[b]{0.24\textwidth}
\centering
\includegraphics[width=3.2cm, height=2.5cm]{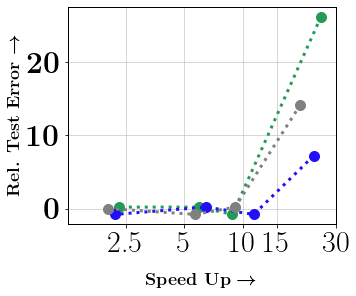}
\caption*{$\underbracket[1pt][1.0mm]{\hspace{3.6cm}}_{\substack{\vspace{-4.0mm}\\
\colorbox{white}{(e) \scriptsize glue-SST2(Random,HB)}}}$}
\phantomcaption
\label{appfig:glue-sst2_random_hb}
\end{subfigure}
\begin{subfigure}[b]{0.24\textwidth}
\centering
\includegraphics[width=3.2cm, height=2.5cm]{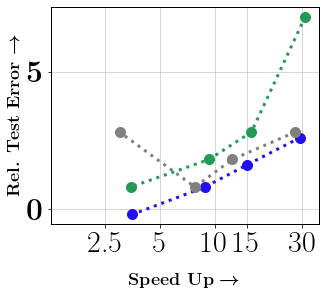}
\caption*{$\underbracket[1pt][1.0mm]{\hspace{3.6cm}}_{\substack{\vspace{-4.0mm}\\
\colorbox{white}{(f) \scriptsize glue-SST2(TPE,HB)}}}$}
\phantomcaption
\label{appfig:glue-sst2_tpe_hb}
\end{subfigure}
\begin{subfigure}[b]{0.24\textwidth}
\centering
\includegraphics[width=3.2cm, height=2.5cm]{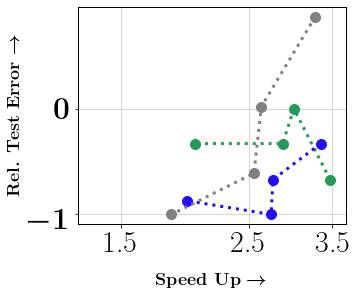}
\caption*{$\underbracket[1pt][1.0mm]{\hspace{3.6cm}}_{\substack{\vspace{-4.0mm}\\
\colorbox{white}{(g) \scriptsize glue-SST2(Random,ASHA)}}}$}
\phantomcaption
\label{appfig:glue-sst2_random_asha}
\end{subfigure}
\begin{subfigure}[b]{0.24\textwidth}
\centering
\includegraphics[width=3.2cm, height=2.5cm]{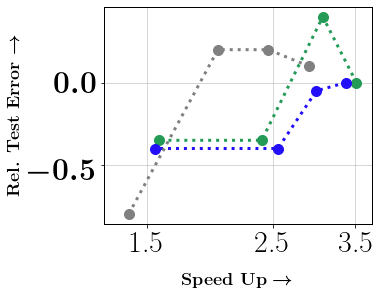}
\caption*{$\underbracket[1pt][1.0mm]{\hspace{3.6cm}}_{\substack{\vspace{-4.0mm}\\
\colorbox{white}{(h) \scriptsize glue-SST2(TPE,ASHA)}}}$}
\phantomcaption
\label{appfig:glue-sst2_tpe_asha}
\end{subfigure}
\begin{subfigure}[b]{0.24\textwidth}
\centering
\includegraphics[width=3.2cm, height=2.5cm]{figs/Test_Accuracy_sst5_random_hb.png}
\caption*{$\underbracket[1pt][1.0mm]{\hspace{3.6cm}}_{\substack{\vspace{-4.0mm}\\
\colorbox{white}{(i) \scriptsize SST5(Random,HB)}}}$}
\phantomcaption
\label{appfig:sst5_random_hb}
\end{subfigure}
\begin{subfigure}[b]{0.24\textwidth}
\centering
\includegraphics[width=3.2cm, height=2.5cm]{figs/Test_Accuracy_sst5_tpe_hb.png}
\caption*{$\underbracket[1pt][1.0mm]{\hspace{3.6cm}}_{\substack{\vspace{-4.0mm}\\
\colorbox{white}{(j) \scriptsize SST5(TPE,HB)}}}$}
\phantomcaption
\label{appfig:sst5_tpe_hb}
\end{subfigure}
\begin{subfigure}[b]{0.24\textwidth}
\centering
\includegraphics[width=3.2cm, height=2.5cm]{figs/Test_Accuracy_sst5_random_asha.png}
\caption*{$\underbracket[1pt][1.0mm]{\hspace{3.6cm}}_{\substack{\vspace{-4.0mm}\\
\colorbox{white}{(k) \scriptsize SST5(Random,ASHA)}}}$}
\phantomcaption
\label{appfig:sst5_random_asha}
\end{subfigure}
\begin{subfigure}[b]{0.24\textwidth}
\centering
\includegraphics[width=3.2cm, height=2.5cm]{figs/Test_Accuracy_sst5_tpe_asha.png}
\caption*{$\underbracket[1pt][1.0mm]{\hspace{3.6cm}}_{\substack{\vspace{-4.0mm}\\
\colorbox{white}{(l) \scriptsize SST5(TPE,ASHA)}}}$}
\phantomcaption
\label{appfig:sst5_tpe_asha}
\end{subfigure}
\begin{subfigure}[b]{0.24\textwidth}
\centering
\includegraphics[width=3.2cm, height=2.5cm]{figs/Test_Accuracy_trec6_random_hb.png}
\caption*{$\underbracket[1pt][1.0mm]{\hspace{3.6cm}}_{\substack{\vspace{-4.0mm}\\
\colorbox{white}{(m) \scriptsize TREC6(Random,HB)}}}$}
\phantomcaption
\label{appfig:trec6_random_hb}
\end{subfigure}
\begin{subfigure}[b]{0.24\textwidth}
\centering
\includegraphics[width=3.2cm, height=2.5cm]{figs/Test_Accuracy_trec6_tpe_hb.png}
\caption*{$\underbracket[1pt][1.0mm]{\hspace{3.6cm}}_{\substack{\vspace{-4.0mm}\\
\colorbox{white}{(n) \scriptsize TREC6(TPE,HB)}}}$}
\phantomcaption
\label{appfig:trec6_tpe_hb}
\end{subfigure}
\begin{subfigure}[b]{0.24\textwidth}
\centering
\includegraphics[width=3.2cm, height=2.5cm]{figs/Test_Accuracy_trec6_random_asha.png}
\caption*{$\underbracket[1pt][1.0mm]{\hspace{3.6cm}}_{\substack{\vspace{-4.0mm}\\
\colorbox{white}{(o) \scriptsize TREC6(Random,ASHA)}}}$}
\phantomcaption
\label{appfig:trec6_random_asha}
\end{subfigure}
\begin{subfigure}[b]{0.24\textwidth}
\centering
\includegraphics[width=3.2cm, height=2.5cm]{figs/Test_Accuracy_trec6_tpe_asha.png}
\caption*{$\underbracket[1pt][1.0mm]{\hspace{3.6cm}}_{\substack{\vspace{-4.0mm}\\
\colorbox{white}{(p) \scriptsize TREC6(TPE,ASHA)}}}$}
\phantomcaption
\label{appfig:trec6_tpe_asha}
\end{subfigure}
\caption{Tuning Results on Text Datasets: Comparison of  performance of \model\ with baselines(\textsc{Random, Craig, Full}) for Hyper-parameter tuning. In sub-figures (a-p), we present speedup {\em vs.} relative test error (in \%), compared to Full data tuning for different methods. On each scatter plot, smaller subsets appear on the right, and larger ones appear on the left. Results are shown for (a-d) SST2, (e-h) glue-SST2, (i-l) SST5, (m-p) TREC6 datasets with different combinations of hyper-parameter search and scheduling algorithms. \emph{The scatter plots show that \model{} achieves the best speedup-accuracy tradeoff in almost every case  \textbf{(bottom-right corner of each plot indicates the best speedup-accuracy tradeoff region})}.}
\label{appfig:text_experiments}
\end{figure}

\begin{figure}[t]
\centering
\includegraphics[width = 14cm, height=0.6cm]{figs/legend.pdf}
\centering
\hspace{-0.6cm}
\begin{subfigure}[b]{0.24\textwidth}
\centering
\includegraphics[width=3.2cm, height=2.5cm]{figs/Test_Accuracy_cifar10_random_hb.png}
\caption*{$\underbracket[1pt][1mm]{\hspace{3.6cm}}_{\substack{\vspace{-4.0mm}\\
\colorbox{white}{(a) \scriptsize CIFAR10(Random,HB)}}}$}
\phantomcaption
\label{appfig:cifar10_random_hb}
\end{subfigure}
\begin{subfigure}[b]{0.24\textwidth}
\centering
\includegraphics[width=3.2cm, height=2.5cm]{figs/Test_Accuracy_cifar10_tpe_hb.png}
\caption*{$\underbracket[1pt][1.0mm]{\hspace{3.6cm}}_{\substack{\vspace{-4.0mm}\\
\colorbox{white}{(b) \scriptsize CIFAR10(TPE,HB)}}}$}
\phantomcaption
\label{appfig:cifar10_tpe_hb}
\end{subfigure}
\begin{subfigure}[b]{0.24\textwidth}
\centering
\includegraphics[width=3.2cm, height=2.5cm]{figs/Test_Accuracy_cifar10_random_asha.png}
\caption*{$\underbracket[1pt][1.0mm]{\hspace{3.6cm}}_{\substack{\vspace{-4.0mm}\\
\colorbox{white}{(c) \scriptsize CIFAR10(Random,ASHA)}}}$}
\phantomcaption
\label{appfig:cifar10_random_asha}
\end{subfigure}
\begin{subfigure}[b]{0.24\textwidth}
\centering
\includegraphics[width=3.2cm, height=2.5cm]{figs/Test_Accuracy_cifar10_tpe_asha.png}
\caption*{$\underbracket[1pt][1.0mm]{\hspace{3.6cm}}_{\substack{\vspace{-4.0mm}\\
\colorbox{white}{(d) \scriptsize CIFAR10(TPE,ASHA)}}}$}
\phantomcaption
\label{appfig:cifar10_tpe_asha}
\end{subfigure}
\begin{subfigure}[b]{0.24\textwidth}
\centering
\includegraphics[width=3.2cm, height=2.5cm]{figs/Test_Accuracy_cifar100_random_hb.png}
\caption*{$\underbracket[1pt][1mm]{\hspace{3.6cm}}_{\substack{\vspace{-4.0mm}\\
\colorbox{white}{(e) \scriptsize CIFAR100(Random,HB)}}}$}
\phantomcaption
\label{appfig:cifar100_random_hb}
\end{subfigure}
\begin{subfigure}[b]{0.24\textwidth}
\centering
\includegraphics[width=3.2cm, height=2.5cm]{figs/Test_Accuracy_cifar100_tpe_hb.png}
\caption*{$\underbracket[1pt][1.0mm]{\hspace{3.6cm}}_{\substack{\vspace{-4.0mm}\\
\colorbox{white}{(f) \scriptsize CIFAR100(TPE,HB)}}}$}
\phantomcaption
\label{appfig:cifar100_tpe_hb}
\end{subfigure}
\begin{subfigure}[b]{0.24\textwidth}
\centering
\includegraphics[width=3.2cm, height=2.5cm]{figs/Test_Accuracy_cifar100_random_asha.png}
\caption*{$\underbracket[1pt][1.0mm]{\hspace{3.6cm}}_{\substack{\vspace{-4.0mm}\\
\colorbox{white}{(g) \scriptsize CIFAR100(Random,ASHA)}}}$}
\phantomcaption
\label{appfig:cifar100_random_asha}
\end{subfigure}
\begin{subfigure}[b]{0.24\textwidth}
\centering
\includegraphics[width=3.2cm, height=2.5cm]{figs/Test_Accuracy_cifar100_tpe_asha.png}
\caption*{$\underbracket[1pt][1.0mm]{\hspace{3.6cm}}_{\substack{\vspace{-4.0mm}\\
\colorbox{white}{(h) \scriptsize CIFAR100(TPE,ASHA)}}}$}
\phantomcaption
\label{appfig:cifar100_tpe_asha}
\end{subfigure}
\begin{subfigure}[b]{0.24\textwidth}
\centering
\includegraphics[width=3.2cm, height=2.5cm]{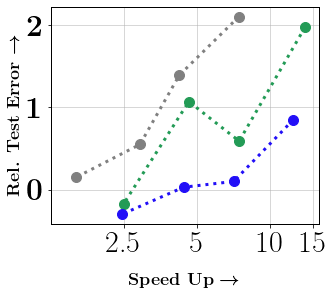}
\caption*{$\underbracket[1pt][1mm]{\hspace{3.6cm}}_{\substack{\vspace{-4.0mm}\\
\colorbox{white}{(i) \scriptsize SVHN(Random,HB)}}}$}
\phantomcaption
\label{appfig:svhn_random_hb}
\end{subfigure}
\begin{subfigure}[b]{0.24\textwidth}
\centering
\includegraphics[width=3.2cm, height=2.5cm]{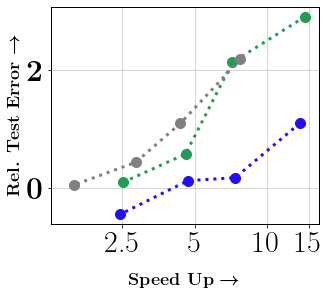}
\caption*{$\underbracket[1pt][1.0mm]{\hspace{3.6cm}}_{\substack{\vspace{-4.0mm}\\
\colorbox{white}{(j) \scriptsize SVHN(TPE,HB)}}}$}
\phantomcaption
\label{appfig:svhn_tpe_hb}
\end{subfigure}
\begin{subfigure}[b]{0.24\textwidth}
\centering
\includegraphics[width=3.2cm, height=2.5cm]{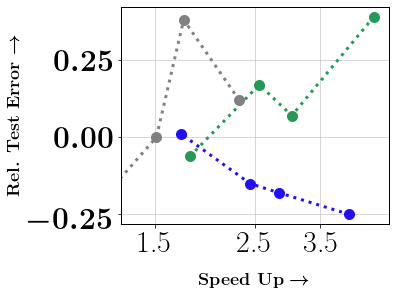}
\caption*{$\underbracket[1pt][1.0mm]{\hspace{3.6cm}}_{\substack{\vspace{-4.0mm}\\
\colorbox{white}{(k) \scriptsize SVHN(Random,ASHA)}}}$}
\phantomcaption
\label{appfig:svhn_random_asha}
\end{subfigure}
\begin{subfigure}[b]{0.24\textwidth}
\centering
\includegraphics[width=3.2cm, height=2.5cm]{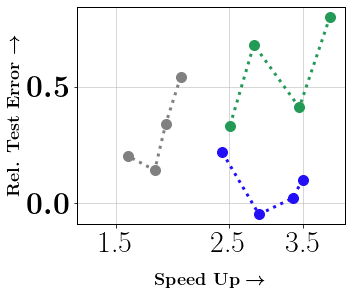}
\caption*{$\underbracket[1pt][1.0mm]{\hspace{3.6cm}}_{\substack{\vspace{-4.0mm}\\
\colorbox{white}{(l) \scriptsize SVHN(TPE,ASHA)}}}$}
\phantomcaption
\label{appfig:svhn_tpe_asha}
\end{subfigure}
\caption{Tuning Results on Image Datasets: Comparison of  performance of \model\ with baselines(\textsc{Random, Craig, Full}) for Hyper-parameter tuning. In sub-figures (a-l), we present speedup {\em vs.} relative test error (in \%), compared to Full data tuning for different methods. On each scatter plot, smaller subsets appear on the right, and larger ones appear on the left. Results are shown for (a-d) CIFAR10, (e-h) CIFAR100, (i-l) SVHN datasets with different combinations of hyper-parameter search and scheduling algorithms. \emph{The scatter plots show that \model{} achieves the best speedup-accuracy tradeoff in almost every case  \textbf{(bottom-right corner of each plot indicates the best speedup-accuracy tradeoff region})}.}
\label{appfig:image_experiments}
\end{figure}

\begin{figure}[t]
\centering
\includegraphics[width = 14cm, height=0.6cm]{figs/legend.pdf}
\centering
\hspace{-0.6cm}
\begin{subfigure}[b]{0.24\textwidth}
\centering
\includegraphics[width=3.2cm, height=2.5cm]{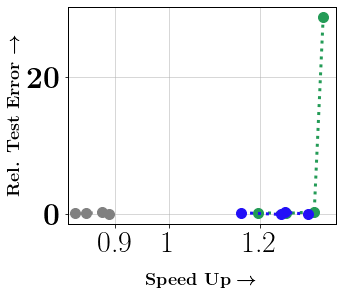}
\caption*{$\underbracket[1pt][1mm]{\hspace{3.6cm}}_{\substack{\vspace{-4.0mm}\\
\colorbox{white}{(a) \scriptsize DNA(Random,HB)}}}$}
\phantomcaption
\label{appfig:dna_random_hb}
\end{subfigure}
\begin{subfigure}[b]{0.24\textwidth}
\centering
\includegraphics[width=3.2cm, height=2.5cm]{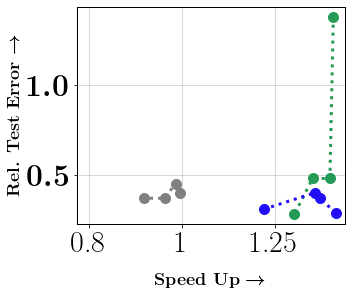}
\caption*{$\underbracket[1pt][1.0mm]{\hspace{3.6cm}}_{\substack{\vspace{-4.0mm}\\
\colorbox{white}{(b) \scriptsize DNA(TPE,HB)}}}$}
\phantomcaption
\label{appfig:dna_tpe_hb}
\end{subfigure}
\begin{subfigure}[b]{0.24\textwidth}
\centering
\includegraphics[width=3.2cm, height=2.5cm]{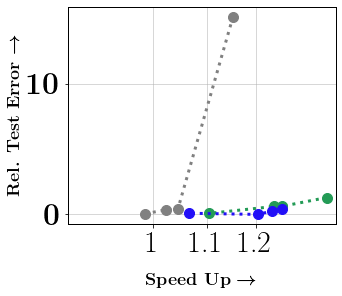}
\caption*{$\underbracket[1pt][1.0mm]{\hspace{3.6cm}}_{\substack{\vspace{-4.0mm}\\
\colorbox{white}{(c) \scriptsize DNA(Random,ASHA)}}}$}
\phantomcaption
\label{appfig:dna_random_asha}
\end{subfigure}
\begin{subfigure}[b]{0.24\textwidth}
\centering
\includegraphics[width=3.2cm, height=2.5cm]{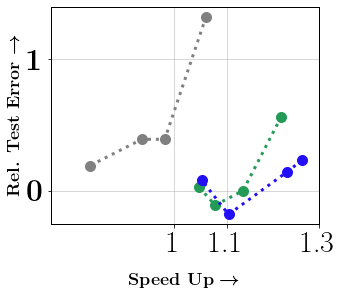}
\caption*{$\underbracket[1pt][1.0mm]{\hspace{3.6cm}}_{\substack{\vspace{-4.0mm}\\
\colorbox{white}{(d) \scriptsize DNA(TPE,ASHA)}}}$}
\phantomcaption
\label{appfig:dna_tpe_asha}
\end{subfigure}
\begin{subfigure}[b]{0.24\textwidth}
\centering
\includegraphics[width=3.2cm, height=2.5cm]{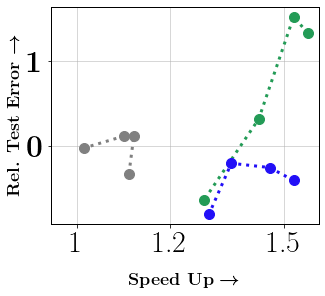}
\caption*{$\underbracket[1pt][1mm]{\hspace{3.6cm}}_{\substack{\vspace{-4.0mm}\\
\colorbox{white}{(e) \scriptsize SATIMAGE(Random,HB)}}}$}
\phantomcaption
\label{appfig:satimage_random_hb}
\end{subfigure}
\begin{subfigure}[b]{0.24\textwidth}
\centering
\includegraphics[width=3.2cm, height=2.5cm]{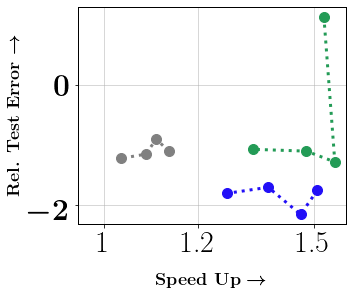}
\caption*{$\underbracket[1pt][1.0mm]{\hspace{3.6cm}}_{\substack{\vspace{-4.0mm}\\
\colorbox{white}{(f) \scriptsize SATIMAGE(TPE,HB)}}}$}
\phantomcaption
\label{appfig:satimage_tpe_hb}
\end{subfigure}
\begin{subfigure}[b]{0.24\textwidth}
\centering
\includegraphics[width=3.2cm, height=2.5cm]{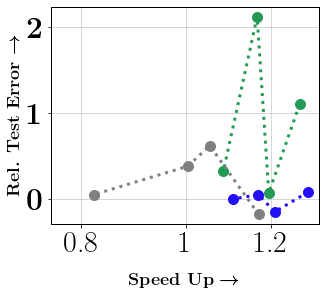}
\caption*{$\underbracket[1pt][1.0mm]{\hspace{3.6cm}}_{\substack{\vspace{-4.0mm}\\
\colorbox{white}{(g) \scriptsize SATIMAGE(Random,ASHA)}}}$}
\phantomcaption
\label{appfig:satimage_random_asha}
\end{subfigure}
\begin{subfigure}[b]{0.24\textwidth}
\centering
\includegraphics[width=3.2cm, height=2.5cm]{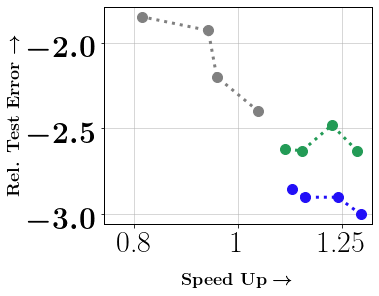}
\caption*{$\underbracket[1pt][1.0mm]{\hspace{3.6cm}}_{\substack{\vspace{-4.0mm}\\
\colorbox{white}{(h) \scriptsize SATIMAGE(TPE,ASHA)}}}$}
\phantomcaption
\label{appfig:satimage_tpe_asha}
\end{subfigure}
\begin{subfigure}[b]{0.24\textwidth}
\centering
\includegraphics[width=3.2cm, height=2.5cm]{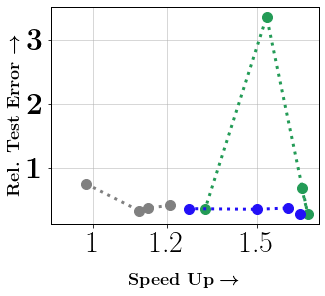}
\caption*{$\underbracket[1pt][1mm]{\hspace{3.6cm}}_{\substack{\vspace{-4.0mm}\\
\colorbox{white}{(i) \scriptsize LETTER(Random,HB)}}}$}
\phantomcaption
\label{appfig:letter_random_hb}
\end{subfigure}
\begin{subfigure}[b]{0.24\textwidth}
\centering
\includegraphics[width=3.2cm, height=2.5cm]{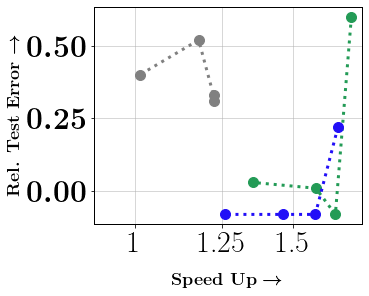}
\caption*{$\underbracket[1pt][1.0mm]{\hspace{3.6cm}}_{\substack{\vspace{-4.0mm}\\
\colorbox{white}{(j) \scriptsize LETTER(TPE,HB)}}}$}
\phantomcaption
\label{appfig:letter_tpe_hb}
\end{subfigure}
\begin{subfigure}[b]{0.24\textwidth}
\centering
\includegraphics[width=3.2cm, height=2.5cm]{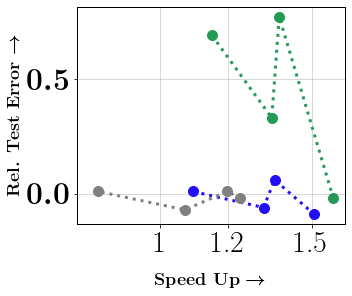}
\caption*{$\underbracket[1pt][1.0mm]{\hspace{3.6cm}}_{\substack{\vspace{-4.0mm}\\
\colorbox{white}{(k) \scriptsize LETTER(Random,ASHA)}}}$}
\phantomcaption
\label{appfig:letter_random_asha}
\end{subfigure}
\begin{subfigure}[b]{0.24\textwidth}
\centering
\includegraphics[width=3.2cm, height=2.5cm]{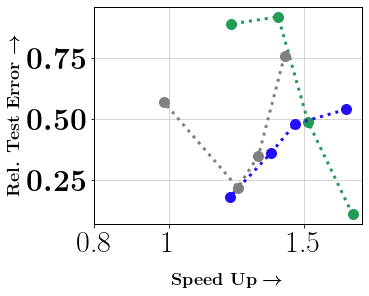}
\caption*{$\underbracket[1pt][1.0mm]{\hspace{3.6cm}}_{\substack{\vspace{-4.0mm}\\
\colorbox{white}{(l) \scriptsize LETTER(TPE,ASHA)}}}$}
\phantomcaption
\label{appfig:letter_tpe_asha}
\end{subfigure}
\begin{subfigure}[b]{0.24\textwidth}
\centering
\includegraphics[width=3.2cm, height=2.5cm]{figs/Test_Accuracy_connect4_random_hb.png}
\caption*{$\underbracket[1pt][1mm]{\hspace{3.6cm}}_{\substack{\vspace{-4.0mm}\\
\colorbox{white}{(m) \scriptsize CONNECT-4(Random,HB)}}}$}
\phantomcaption
\label{appfig:connect4_random_hb}
\end{subfigure}
\begin{subfigure}[b]{0.24\textwidth}
\centering
\includegraphics[width=3.2cm, height=2.5cm]{figs/Test_Accuracy_connect4_tpe_hb.png}
\caption*{$\underbracket[1pt][1.0mm]{\hspace{3.6cm}}_{\substack{\vspace{-4.0mm}\\
\colorbox{white}{(n) \scriptsize CONNECT-4(TPE,HB)}}}$}
\phantomcaption
\label{appfig:connect4_tpe_hb}
\end{subfigure}
\begin{subfigure}[b]{0.24\textwidth}
\centering
\includegraphics[width=3.2cm, height=2.5cm]{figs/Test_Accuracy_connect4_random_asha.png}
\caption*{$\underbracket[1pt][1.0mm]{\hspace{3.6cm}}_{\substack{\vspace{-4.0mm}\\
\colorbox{white}{(o) \scriptsize CONNECT-4(Random,ASHA)}}}$}
\phantomcaption
\label{appfig:connect4_random_asha}
\end{subfigure}
\begin{subfigure}[b]{0.24\textwidth}
\centering
\includegraphics[width=3.2cm, height=2.5cm]{figs/Test_Accuracy_connect4_tpe_asha.png}
\caption*{$\underbracket[1pt][1.0mm]{\hspace{3.6cm}}_{\substack{\vspace{-4.0mm}\\
\colorbox{white}{(p) \scriptsize CONNECT-4(TPE,ASHA)}}}$}
\phantomcaption
\label{appfig:connect4_tpe_asha}
\end{subfigure}
\caption{Tuning Results on Tabular Datasets: Comparison of  performance of \model\ with baselines(\textsc{Random, Craig, Full}) for Hyper-parameter tuning. In sub-figures (a-p), we present speedup {\em vs.} relative test error (in \%), compared to Full data tuning for different methods. On each scatter plot, smaller subsets appear on the right, and larger ones appear on the left. Results are shown for (a-d) DNA, (e-h) SATIMAGE, (i-l) LETTER, (m-p) CONNECT-4 datasets with different combinations of hyper-parameter search and scheduling algorithms. \emph{The scatter plots show that \model{} achieves the best speedup-accuracy tradeoff in almost every case  \textbf{(bottom-right corner of each plot indicates the best speedup-accuracy tradeoff region})}.}
\label{appfig:tabular_experiments}
\end{figure}

We present more hyper-parameter tuning results of \model{} on additional text, image, and tabular datasets in Figures~\ref{appfig:text_experiments},\ref{appfig:image_experiments},\ref{appfig:tabular_experiments}. From the results, it is evident that \model\ achieves best speedup vs. accuracy tradeoff in almost all of the cases.


\subsection{CO2 Emissions and Energy Consumption Results} \label{app:co2emissions}
\begin{figure}[t]
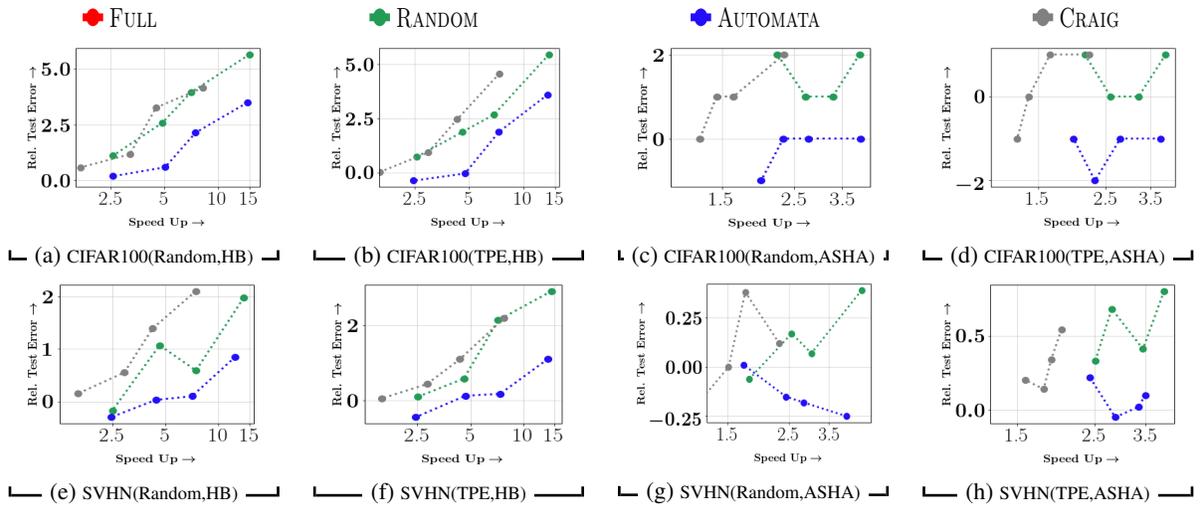

\centering
\includegraphics[width = 14cm, height=0.6cm]{figs/legend.pdf}
\centering
\hspace{-0.6cm}
\begin{subfigure}[b]{0.24\textwidth}
\centering
\includegraphics[width=3.2cm, height=2.5cm]{figs/Test_Accuracy_cifar100_random_hb.png}
\caption*{$\underbracket[1pt][1mm]{\hspace{3.6cm}}_{\substack{\vspace{-4.0mm}\\
\colorbox{white}{(a) \scriptsize CIFAR100(Random,HB)}}}$}
\phantomcaption
\label{appfig:cifar100_energy_random_hb}
\end{subfigure}
\begin{subfigure}[b]{0.24\textwidth}
\centering
\includegraphics[width=3.2cm, height=2.5cm]{figs/Test_Accuracy_cifar100_tpe_hb.png}
\caption*{$\underbracket[1pt][1.0mm]{\hspace{3.6cm}}_{\substack{\vspace{-4.0mm}\\
\colorbox{white}{(b) \scriptsize CIFAR100(TPE,HB)}}}$}
\phantomcaption
\label{appfig:cifar100_energy_tpe_hb}
\end{subfigure}
\begin{subfigure}[b]{0.24\textwidth}
\centering
\includegraphics[width=3.2cm, height=2.5cm]{figs/Test_Accuracy_cifar100_random_asha.png}
\caption*{$\underbracket[1pt][1.0mm]{\hspace{3.6cm}}_{\substack{\vspace{-4.0mm}\\
\colorbox{white}{(c) \scriptsize CIFAR100(Random,ASHA)}}}$}
\phantomcaption
\label{appfig:cifar100_energy_random_asha}
\end{subfigure}
\begin{subfigure}[b]{0.24\textwidth}
\centering
\includegraphics[width=3.2cm, height=2.5cm]{figs/Test_Accuracy_cifar100_tpe_asha.png}
\caption*{$\underbracket[1pt][1.0mm]{\hspace{3.6cm}}_{\substack{\vspace{-4.0mm}\\
\colorbox{white}{(d) \scriptsize CIFAR100(TPE,ASHA)}}}$}
\phantomcaption
\label{appfig:cifar100_energy_tpe_asha}
\end{subfigure}
\begin{subfigure}[b]{0.24\textwidth}
\centering
\includegraphics[width=3.2cm, height=2.5cm]{figs/Test_Accuracy_svhn_random_hb.png}
\caption*{$\underbracket[1pt][1mm]{\hspace{3.6cm}}_{\substack{\vspace{-4.0mm}\\
\colorbox{white}{(e) \scriptsize SVHN(Random,HB)}}}$}
\phantomcaption
\label{appfig:svhn_co2_random_hb}
\end{subfigure}
\begin{subfigure}[b]{0.24\textwidth}
\centering
\includegraphics[width=3.2cm, height=2.5cm]{figs/Test_Accuracy_svhn_tpe_hb.png}
\caption*{$\underbracket[1pt][1.0mm]{\hspace{3.6cm}}_{\substack{\vspace{-4.0mm}\\
\colorbox{white}{(f) \scriptsize SVHN(TPE,HB)}}}$}
\phantomcaption
\label{appfig:svhn_co2_tpe_hb}
\end{subfigure}
\begin{subfigure}[b]{0.24\textwidth}
\centering
\includegraphics[width=3.2cm, height=2.5cm]{figs/Test_Accuracy_svhn_random_asha.png}
\caption*{$\underbracket[1pt][1.0mm]{\hspace{3.6cm}}_{\substack{\vspace{-4.0mm}\\
\colorbox{white}{(g) \scriptsize SVHN(Random,ASHA)}}}$}
\phantomcaption
\label{appfig:svhn_co2_random_asha}
\end{subfigure}
\begin{subfigure}[b]{0.24\textwidth}
\centering
\includegraphics[width=3.2cm, height=2.5cm]{figs/Test_Accuracy_svhn_tpe_asha.png}
\caption*{$\underbracket[1pt][1.0mm]{\hspace{3.6cm}}_{\substack{\vspace{-4.0mm}\\
\colorbox{white}{(h) \scriptsize SVHN(TPE,ASHA)}}}$}
\phantomcaption
\label{appfig:svhn_co2_tpe_asha}
\end{subfigure}
\caption{Comparison of  performance of \model\ with baselines(\textsc{Random, Craig, Full}) for Hyper-parameter tuning. In sub-figures (a-d), we present energy ratio {\em vs.} relative test error (in \%), compared to Full data tuning for different methods on CIFAR100 dataset. In sub-figures (e-h), we present co2 emissions ratio {\em vs.} relative test error (in \%), compared to Full data tuning for different methods on SVHN dataset. On each scatter plot, smaller subsets appear on the right, and larger ones appear on the left. \emph{The scatter plots show that \model{} achieves the best energy savings and CO2 reductions, thereby achieving the best efficiency vs. performance tradeoff in almost every case. \textbf{(Bottom-right corner of each plot indicates the best efficiency vs. performance tradeoff region)}}.}
\label{appfig:energy_co2}
\end{figure}

Sub-figures~\ref{appfig:cifar100_energy_random_hb},\ref{appfig:cifar100_energy_tpe_hb},\ref{appfig:cifar100_energy_random_asha},\ref{appfig:cifar100_energy_tpe_asha} shows the energy efficiency plot of \model{} on CIFAR100 dataset for 1\%, 5\%, 10\%, 30\% subset fractions. For calculating the energy consumed by the GPU/CPU cores, we use pyJoules\footnote{\scriptsize{\url{https://pypi.org/project/pyJoules/}}.}. From the plot, it is evident that \model{} is more energy efficient compared to the other baselines and full data tuning. Sub-figures~\ref{appfig:svhn_co2_random_hb},\ref{appfig:svhn_co2_tpe_hb},\ref{appfig:svhn_co2_random_asha},\ref{appfig:svhn_co2_tpe_asha} shows the plot of relative error vs CO2 emissions efficiency, both w.r.t full training. CO2 emissions were estimated based on the total compute time using the \href{https://mlco2.github.io/impact#compute}{Machine Learning Impact calculator} presented in \cite{lacoste2019quantifying}. From the results, it is evident that \model\ achieved the best energy vs. accuracy tradeoff and is environmentally friendly based on CO2 emissions compared to other baselines (including \textsc{Craig} and \textsc{Random}).

\end{document}